\documentclass{acmsiggraph}

\TOGonlineid{0262}


\CopyrightYear{2017}
\setcopyright{acmcopyright}
\conferenceinfo{CONFERENCE PROGRAM NAME}{MONTH, DAY, and YEAR} 
\isbn{THIS-IS-A-SAMPLE-ISBN}\acmPrice{\$15.00}
\doi{http://doi.acm.org/THIS/IS/A/SAMPLE/DOI}

\usepackage{amsfonts}       
\usepackage{amsmath}
\usepackage{booktabs}       
\usepackage{enumitem}
\usepackage{bm}
\usepackage{nicefrac}       
\usepackage{subcaption}
\usepackage{adjustbox}

\DeclareMathOperator{\E}{\mathbb{E}}

\newcommand{\etal}{\textit{et al}.}
\newcommand{\ie}{\textit{i}.\textit{e}.}
\newcommand{\eg}{\textit{e}.\textit{g}.}

\title{Mastering Sketching:\\Adversarial Augmentation for Structured Prediction}

\author{Edgar Simo-Serra\thanks{The authors assert equal contribution and joint first authorship.} \and
        Satoshi Iizuka\footnotemark[1]\hspace{5pt} \and
        Hiroshi Ishikawa }
\affiliation{Waseda University}
\pdfauthor{Edgar Simo-Serra, Satoshi Iizuka, and Hiroshi Ishikawa}

\begin{document}

\teaser{
  \includegraphics[width=\linewidth]{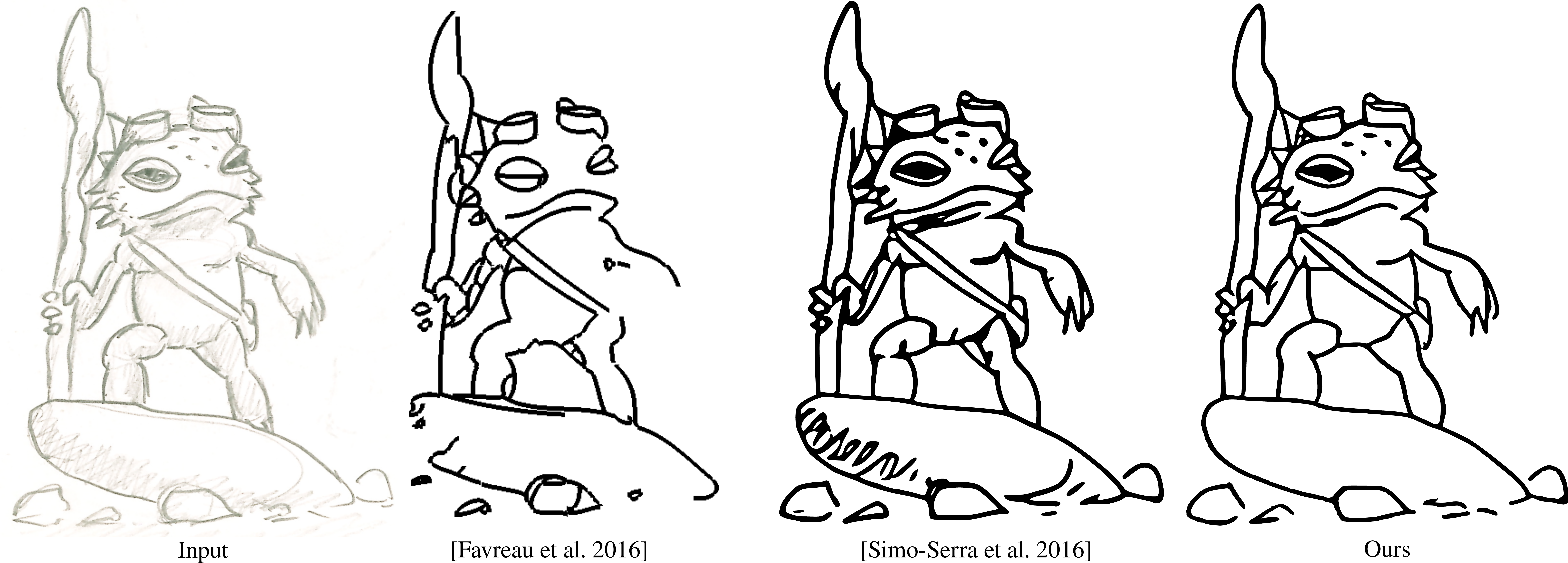}
  \vspace{-5mm}
  \caption{Comparison of our approach for sketch simplification with the state
  of the art. We note that the approach of [Favreau et al. 2016] required
  significant parameter tuning in order to obtain the results shown above.
  Existing approaches miss important lines and conserve superfluous details
  such as shading and support scribbles, which hampers further processing of
  the drawing, \eg, coloring.}
  \label{fig:teaser}
}

\maketitle

 
\begin{abstract}
   We present an integral framework for training sketch simplification networks
   that convert challenging rough sketches into clean line drawings.
   Our approach augments a simplification network with a discriminator network,
   training both networks jointly so that the discriminator network discerns whether a line drawing 
   is a real training data or the output of the simplification network, which in turn tries to fool it.
   This approach has two major advantages. First, because the discriminator network
   learns the structure in line drawings, it encourages the output sketches of the
   simplification network to be more similar in appearance to the training sketches.
   Second, we can also train the networks with additional unsupervised data:
   by adding rough sketches and line drawings that are not corresponding to each other,
   we can improve the quality of the sketch simplification.
   Thanks to a difference in the architecture, our approach has advantages over similar
   adversarial training approaches in stability of training and the aforementioned ability to
   utilize unsupervised training data.
   We show how our framework can be used to train models that significantly outperform
   the state of the art in the sketch simplification task, despite using the same architecture 
   for inference.
   We additionally present an approach to optimize for a single image, which improves 
   accuracy at the cost of additional computation time.
   Finally, we show that, using the same framework, it is possible to train the
   network to perform the inverse problem, \ie, convert simple line sketches into pencil 
   drawings, which is not possible using the standard mean squared error loss. We
   validate our framework with two user tests, where our approach is preferred
   to the state of the art in sketch simplification 92.3\% of the time and
   obtains 1.2 more points on a scale of 1 to 5.
\end{abstract}


%
%


\keywordlist

\conceptlist


\begin{figure}[th]
\includegraphics[width=\linewidth]{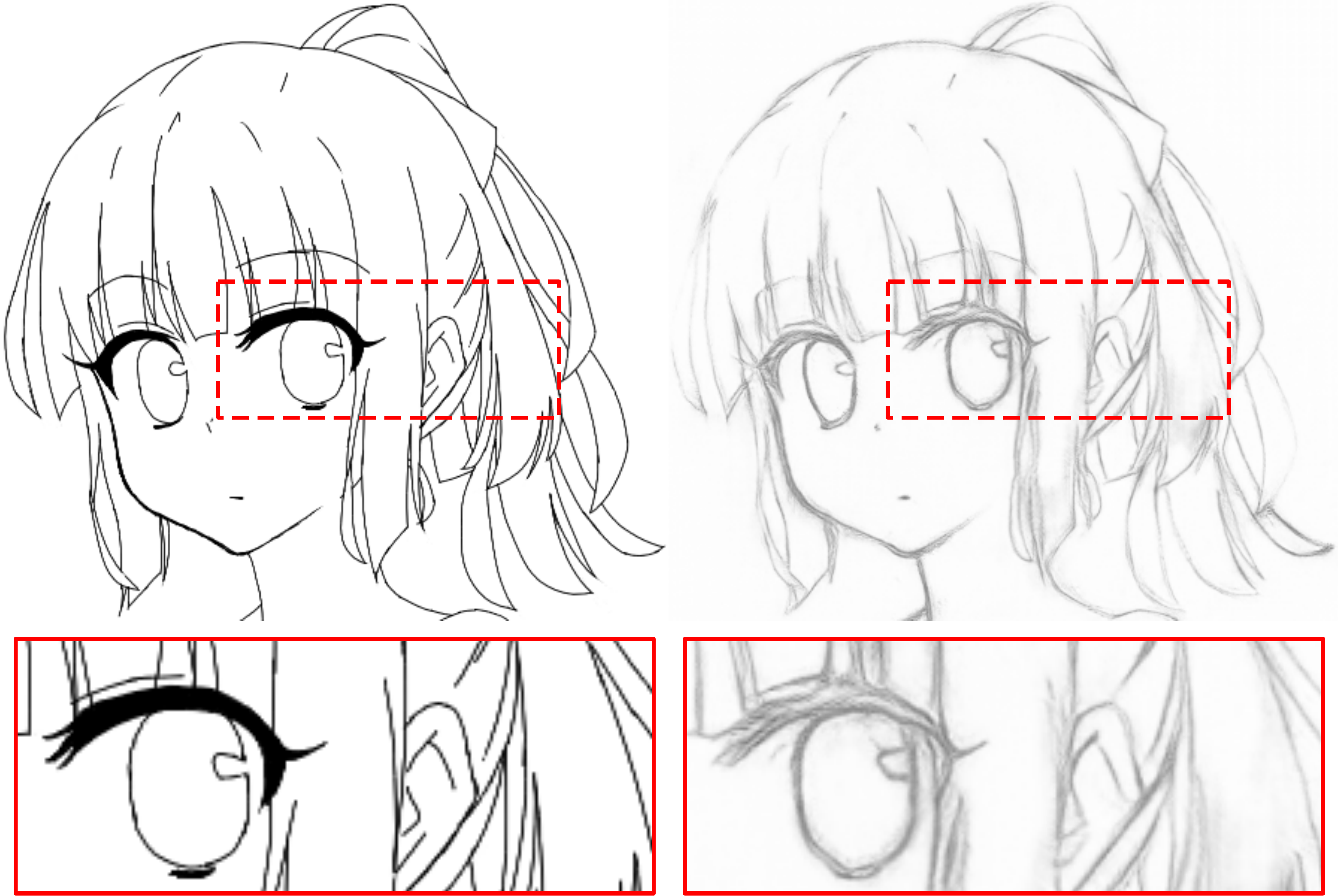} 
\vspace{-5mm}
\caption{Pencil drawing generation with our framework. The line drawing on the left is turned into the pencil drawing on the right.}
\label{fig:teaser_rough}
\end{figure}

\section{Introduction}

\begin{figure*}[t]
\begin{center}
\begin{tabular}{cc}
Annotated Images &
Sketches ``in the wild'' \\
\includegraphics[height=6cm]{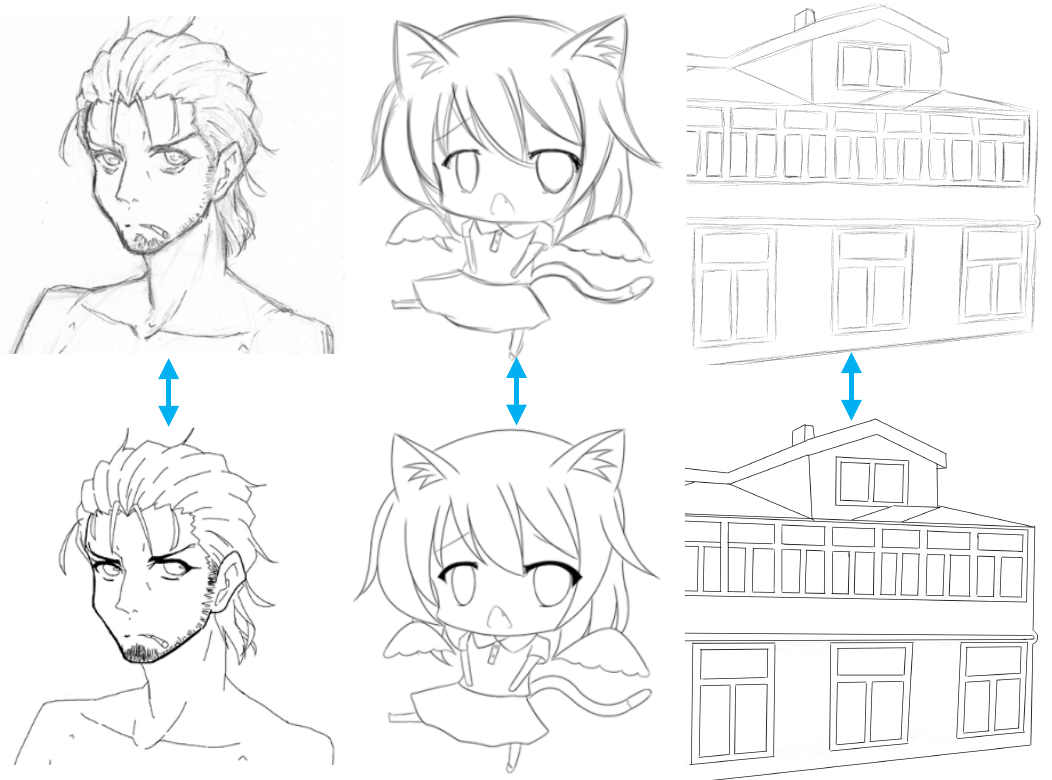} &
\includegraphics[height=6cm]{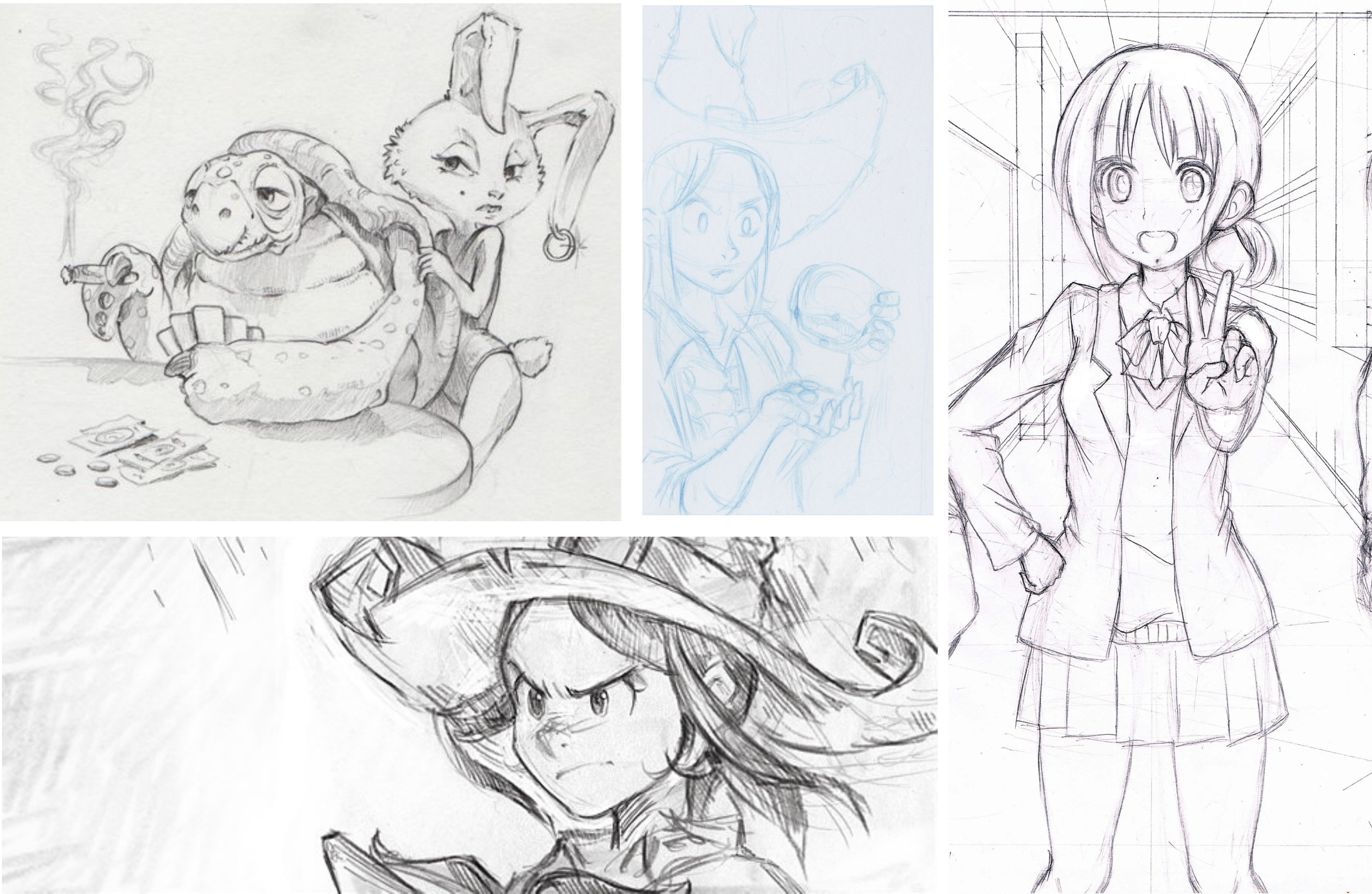} \\
\end{tabular}
\end{center}
\vspace{-3mm}
\caption{Comparison between the supervised dataset of [Simo-Serra
et al. 2016] and rough sketches found in the wild. The difficulty of obtaining
high quality and diverse rough sketches and their corresponding simplified
sketches greatly limits performance on rough sketches ``in the wild'' that can
be significantly different from the annotated data used for training models.
The three images on the left of the Sketches ``in the wild'' are copyrighted by
\href{http://www.davidrevoy.com}{David Revoy www.davidrevoy.com} and licensed
under CC-by 4.0.
}
\label{fig:difficulty}
\end{figure*}

Sketching plays a critical role in the initial stages of artistic work such as
product design and animation, allowing an artist to quickly draft and visualize
their thoughts. However, the process of digitizing and cleaning up the rough
pencil drawings involves a large overhead. This process is called
\textit{sketch simplification}, and involves simplifying multiple overlapped
lines into a single line and erasing superfluous lines that are used as
references when drawing, as shown in Fig.~\ref{fig:teaser}. Due to the large
variety of drawing styles and tools, developing a generic approach for sketch
simplification is a challenging task.  In this work, we propose a novel
approach for sketch simplification that is able to outperform current
approaches. Furthermore, our proposed framework can also be used to do the
inverse problem, \ie, pencil drawing generation as shown in
Fig.~\ref{fig:teaser_rough}.

Recently, an approach to automatize the sketch simplification task with a fully
convolutional network was proposed by Simo-Serra
\etal~\shortcite{SimoSerraSIGGRAPH2016}. In order to train this network, large
amounts of supervised data, consisting of pairs of rough sketches and their
corresponding sketch simplifications, was obtained. To collect this data,
the tedious process of inverse dataset construction was used. This involves
hiring artists to draw a rough sketch, simplify the sketch into a clean line
drawing, and finally redraw the rough sketch on top of the line drawing to
ensure the training data is aligned. This is not only time- and money- consuming,
but the resulting data also differs from the true rough sketches drawn without clean
line drawings as references, as shown in Fig.~\ref{fig:difficulty}. The
resulting models trained with this data therefore generalize poorly to rough sketches
found ``in the wild'', which are representative of true sketch simplification
usage cases. Here, we propose a framework that can incorporate these unlabeled
sketches found ``in the wild'' into the learning process and significantly improve the
performance in real usage cases.

Our approach combines a
fully-convolutional sketch simplification network with a discriminator network
that is trained to discern real line drawings from those generated by a
network. The simplification network is trained to both simplify sketches, and
\textit{also} to deceive the discriminator network.  In contrast to optimizing
with only the standard mean squared error loss, which only considers individual
pixel and not their neighbors, our proposed loss considers the entire output
image. This allows significantly improving the quality of the sketch
simplifications. Furthermore, the same framework allows augmenting the
supervised training dataset with unsupervised examples, leading to a hybrid
supervised and unsupervised learning framework which we call
\textit{Adversarial Augmentation}.  We evaluate our approach on a diverse set
of challenging rough sketches, comparing against the state of the art and
alternate optimization strategies. We also perform a user study, in
which our approach is preferred 92.3\% of the time to the leading competing approach.

We also evaluate our framework on the inverse sketch simplification problem,
\ie, generating pencil drawings from line drawings. We note that, unlike the
sketch simplification problem, converting clean sketches into rough sketches
cannot be accomplished by using a straightforward supervised approach
such as a mean squared error loss: the model is unable to learn the structure
of the output, and instead of producing rough pencil drawings, it produces
a blurry representations of the input. However, by using our adversarial augmentation
framework, we can successfully train a model to convert clean sketches into
rough sketches.

Finally, as another usage case of our framework, we also consider the case of
augmenting the training data with the test data input. We note that we use the
test data in an unsupervised manner: no privileged information is used. By
using the discriminator network and training jointly with the test images, we
can significantly improve the results at the cost of computational efficiency.

\noindent \textbf{In summary}, we present:
\vspace{-2mm}
\begin{itemize}[noitemsep,nolistsep,leftmargin=*]
\item A unified framework to jointly train from supervised and unsupervised data.
\item Significant improvements over the state of the art in sketch simplification quality.
\item A method to further improve the quality by single image training.
\item The first approach to converting simplified line drawings into rough pencil-drawn-like images.
\end{itemize}

\section{Background}

\paragraph{Sketch Simplification.}
As sketch simplification is a tedious manual task for most artists,
many approaches attempting to automate it have been proposed. 
A common approach is assisting the user by adjusting the strokes by,
for instance, using geometric constraints~\cite{Igarashi:1997:IBT:263407.263525}, fitting
B{\'e}zier curves~\cite{Bae:2008:IAS:1449715.1449740}, or merging strokes based
on heuristics~\cite{Grimm:2012:JDS:2331067.2331084}. These approaches require
all the strokes and their drawing order as input.
Along similar lines, many work has focused on the problem of simplifying vector
images~\cite{CGF:CGF1151,5710858,LindlbauerHHSS2013,FiserSketch2015,LiuSIGGRAPH2015}.
However, approaches that can be used on raster images have been unable to process
complex real-world sketches~\cite{NorisTOG2013,ChenCGF2013}. Most of these
approaches rely on a pre-processing stage that converts the image into
a graph which is then optimized~\cite{HilairePAMI2006,NorisTOG2013,FavreauSIGGRAPH2016};
however, they generally cannot recover from errors in the pre-processing stage. 
Simo-Serra \etal~\shortcite{SimoSerraSIGGRAPH2016} proposed a
fully-automatic approach for simplifying sketches directly from raster images
of rough sketches by using a fully-convolutional network. 
However, their approach still needs large amounts of supervised data,
consisting of pairs of rough sketches and their corresponding sketch
simplifications, tediously created by the process of inverse dataset
construction. Their training sketches differ from the true rough sketches, as
shown in Fig.~\ref{fig:difficulty}.  The resulting models trained with this
data therefore generalize poorly to real rough sketches.  We build upon their
work and propose a framework that can incorporate unlabeled real sketches into
the learning process and significantly improve the performance in real usage
cases.  Thus, in contrast to all previous works, we consider challenging
real-world scanned data that is significantly more complex than previously
attempted. 


\paragraph{Generative Adversarial Networks.}
In order to train generative models using unsupervised data with back-propagation,
Goodfellow \etal~\shortcite{GoodfellowNIPS2014} proposed the Generative Adversarial
Networks (GAN).
In the GAN approach, a generative model is paired with a
discriminative model and trained jointly. The discriminative model is trained
to discern whether or not an image is real or artificially generated, while the
generative model is trained to deceive the discriminative model. By training
both jointly, it is possible to train the generative model to create realistic
images from random inputs~\cite{RadfordICLR2016}.
There is also a variant, Conditional GAN (CGAN),  that learns a conditional generative model.
This can be used to generate images conditioned on class labels~\cite{MirzaNIPSW2014}.
In a concurrent work, using CGAN for the image-to-image synthesis 
problem was recently proposed in a preprint~\cite{IsolaARXIV2016}, where the authors use
a CGAN loss and apply it to tasks such as image colorization and scene reconstruction from labels.
However, CGAN is unable to use unsupervised data, which helps improve
performance significantly. In this paper, we compare against a strong CGAN-baseline, 
using the sketch simplification model of~\cite{SimoSerraSIGGRAPH2016} with 
large amounts of data augmentation, and show that our approach can generate 
significantly better sketch simplification.
The difference in architecture of our approach compared with CGAN is illustrated
in Fig.~\ref{fig:trainingloss}.

\paragraph{Pencil Drawing Generation.}
To our knowledge, the inverse problem of converting clean sketches to pencil
drawings has not been tackled before. Making natural images appear like
sketches has been widely
studied~\cite{KangNPAR2007,LuNPAR2012}, as natural images have
rich gradients which can be exploited for the task. However, using clean sketches that
contain very sparse information as inputs is an entirely different problem. In
order to produce realistic strokes, most approaches rely on a dataset of
examples~\cite{BergerSIGGRAPH2013}, while our approach can directly create
novel realistic rough-sketch strokes. As we will show, the discriminative
adversarial training proves critical in obtaining realistic pencil-drawn
outputs.



\subsection{Deep Learning}
We base our work on deep multi-layer convolutional neural
networks~\cite{FukushimaNN1988,LecunNM1989}, which have seen a surge in usage
in the past few years, and have seen application in a wide variety of problems.
Just restricting our attention to those with image input and output, there are such recent work as
super-resolution~\cite{DongPAMI2016}, semantic segmentation~\cite{NohICCV2015}, and image
colorization~\cite{IizukaSIGGRAPH2016}. These networks are all built upon
convolutional layers of the form:
\begin{equation}%
y^c_{u,v} = \sigma\left( \sum_k \left( b^{c,k} + \sum_{i=u-M}^{u+M} \sum_{j=v-N}^{v+N} w^{c,k}_{i+M,j+N} \, x^k_{i,j} \right) \right),
\end{equation}%
\noindent where for a $(2M+1)\times(2N+1)$ convolution kernel, each output
channel $c$ and coordinates $(u,v)$, the output value $y^c_{u,v}$ is computed as
an affine transformation of the input pixel $x^k_{u,v}$ for all input channels
$k$ with a shared weight matrix formed by $w^{c,k}$ and bias value $b^{c,k}$ that is
run through a non-linear activation function $\sigma(\cdot)$. The most widely
used non-linear activation function is the Rectified Linear Unit (ReLU) where
$\sigma(x)=\max(0,x)$~\cite{NairICML2010}.

These layers are a series of learnable filters with $w$ and
$b$ being the learnable parameters. In order to train a network, a dataset
consisting of pairs of input and their corresponding ground truth $(x,y^*)$ are used in
conjunction with a loss function $L(y,y^*)$ that measures the error between the
output $y$ of the network and the ground truth $y^*$. This error is
used to update the learnable parameters with the backpropagation
algorithm~\cite{RumelhartNATURE1986}. In this work we also consider the
scenario in which not all data is necessarily in the form of pairs $(x,y^*)$,
but can also be in the form of single samples $x$ and $y^*$ that are not
corresponding pairs.

Our work is based on fully-convolutional neural network models that can be
applied to images of any resolution. These networks generally follow an
encoder-decoder architecture, in which the first layers of the network have
an increased stride to lower the resolution of the input layer. At lower
resolutions, the subsequent layers are able to process larger regions of the
input image: for instance, a $3\times3$-pixel convolution on an image at half resolution
is computed with a $5\times5$-pixel area of the original image. Additionally, by
processing at lower resolutions, both the memory requirements and computation
times are significantly decreased. 
In this paper, we base our network model on that
of~\cite{SimoSerraSIGGRAPH2016} and show that we can greatly improve the 
performance of the resulting model by using a significantly improved learning approach.

\section{Adversarial Augmentation}

\begin{figure}[t]
   \includegraphics[width=\linewidth]{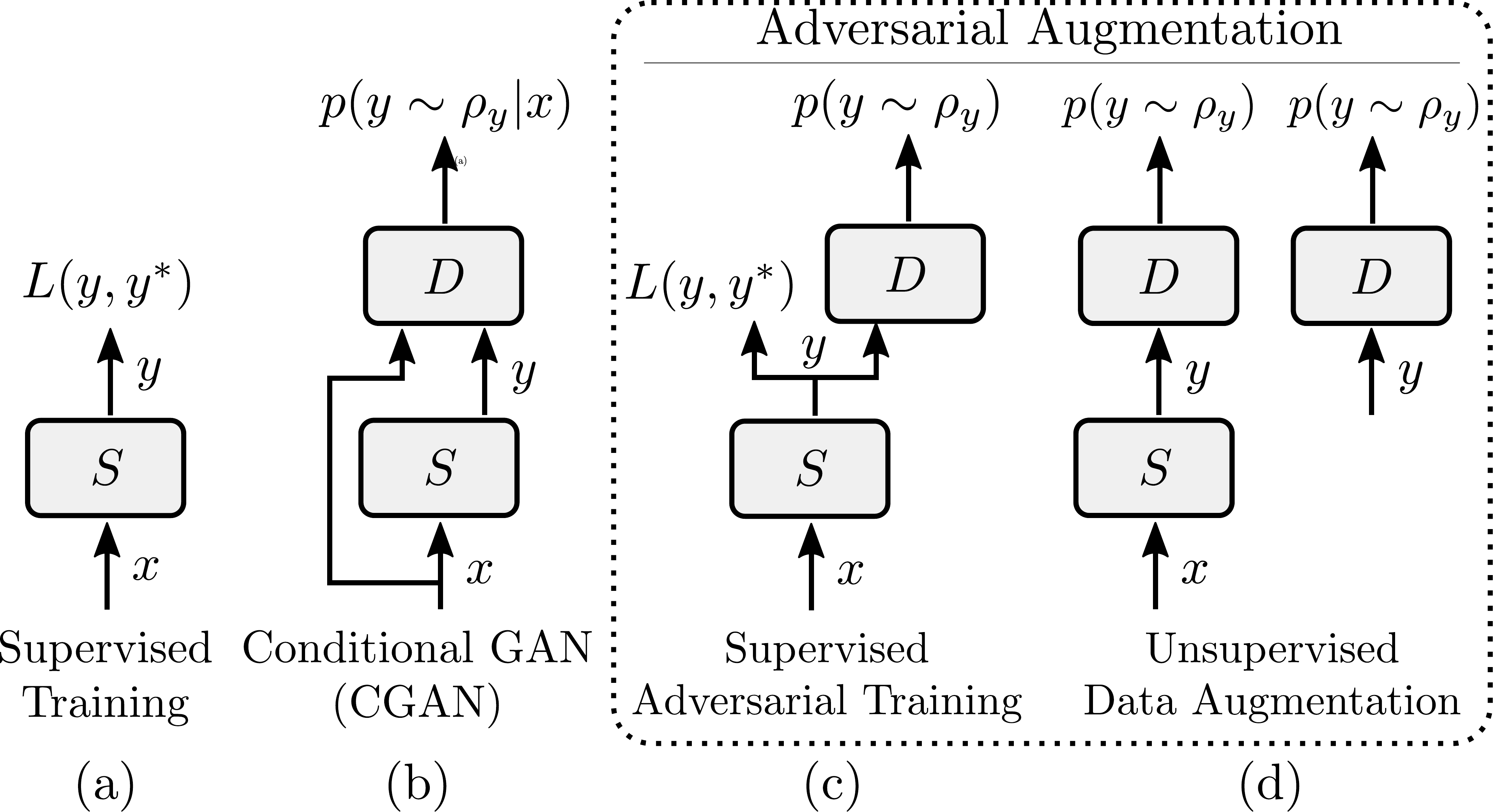} \vspace{-4mm}%
   \caption{Overview of our approach, Adversarial Augmentation, compared with different 
   network training approaches.
   (a) To train a prediction network $S: x\mapsto y$, the supervised training trains by 
   using a specific loss $L(y,y^*)$ that encourages the output $y$ of the network to
   match that of the ground truth data $y^*$. 
   (b) The Conditional Generative Adversarial Network (CGAN) introduces an additional
   discriminator network $D$ that attempts to discern whether or not an image is from the training data 
   or a prediction by the prediction network $S$, while $S$ is trained to deceive $D$.
   The discriminator network $D$ takes two inputs $x$ and $y$ and estimates the conditional probability
   that  $y$ comes from the training data given $x$. 
   (c) Our approach fuses the supervised training and the adversarial training. 
   We use a specific loss $L(y,y^*)$ to force the output to be coherent with the input, while 
   also using a discriminator network $D$, similar to CGAN, but not conditioned on $x$
   (Supervised Adversarial Training).
   (d) The fact that $S$ only takes $x$ as input, and that $D$ only takes $y$, enables us to use training
   data $x$ and $y$ that are not related, \ie, in an unsupervised manner, to further train $S$ and $D$
   (Unsupervised Data Augmentation).}
   \label{fig:trainingloss}
\end{figure}

\begin{figure}[t]
\includegraphics[width=\linewidth]{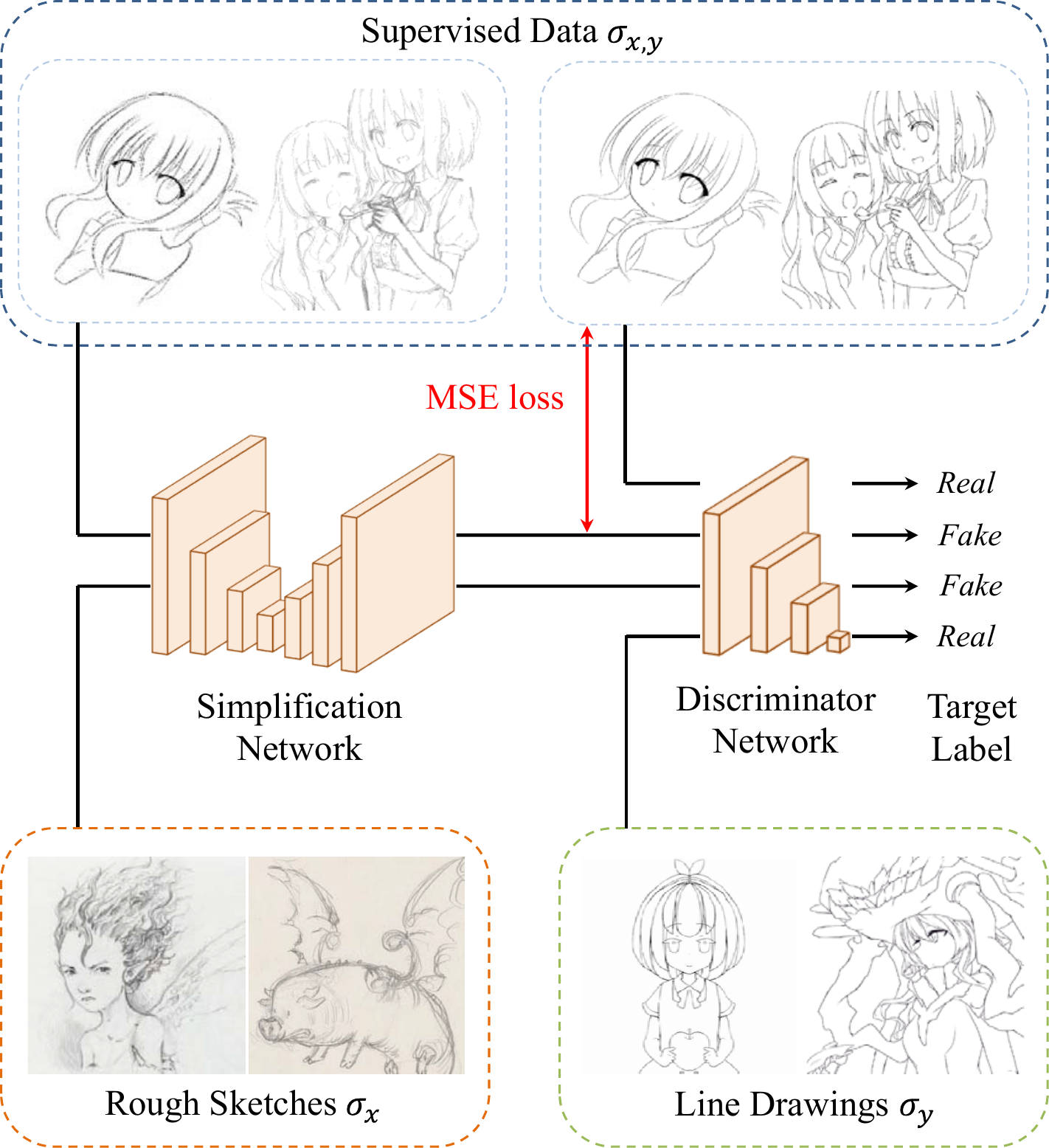}
\vspace{-5mm}
\caption{Overview of our adversarial augmentation framework. We train the
simplification network using both supervised pairs of data $\rho_{x,y}$, and
unsupervised data $\rho_x$ and $\rho_y$. The discriminator network is trained
to distinguish real line drawings from those output by the simplification
network, while the simplification network is trained to deceive the
discriminator network. For the data pairs we additionally use the MSE loss
which forces the simplification network outputs to correspond to the input
rough sketches.
The two images forming Rough Sketches $\sigma_x$ are copyrighted by
\href{http://www.davidrevoy.com}{David Revoy www.davidrevoy.com} and licensed
under CC-by 4.0.
}
\label{fig:overview2}
\end{figure}

We present \textit{adversarial augmentation}, which is the fusion of unsupervised
and adversarial training 
focused on the purpose of augmenting existing networks for structured prediction tasks.
An overview of our approach compared with different training approaches can be seen in
Fig.~\ref{fig:trainingloss}. 
Similar to Generative Adversarial Networks (GAN)~\cite{GoodfellowNIPS2014}, we employ a discriminator network that
attempts to distinguish whether an image comes from real data or is the
output of another network. Unlike in the case of standard supervised losses such as the Mean
Squared Error (MSE), with the discriminator network the
output is encouraged to have a global consistency similar to the training
images. Furthermore, in contrast to other existing approaches, our proposed
approach permits augmenting the training dataset with unlabeled examples, both
increasing performance and generalization, while having much more stable training.
An overview of our framework can be seen in Fig.~\ref{fig:overview2}.

While this work focuses on sketch simplification, the presented approach is
general and applicable to other structured prediction problems, such as semantic segmentation
or saliency detection.

\subsection{The GAN Framework}
The purpose of Generative Adversarial Network (GAN)~\cite{GoodfellowNIPS2014}
is, given a training set of samples, estimating a generative model that
stochastically generates samples similar to those drawn from a distribution represented by the
given training set $\rho_y$. So a trained generative model produces, for
instance, pictures of random cars similar to given set of sample pictures. A
generative model is described as a neural network model $G:z \mapsto y$ that
maps a random vector $z$ to an output $y$. In the GAN framework, we train two
network models: a) the generative model $G$ above, and b) a discriminator model
$D:y \mapsto D(y)\in \mathbb{R}$ that computes the probability that a
structured input (\eg, image) $y$ came from the real data, rather than the
output of $G$. We jointly optimize $G$ and $D$ with respect to the expectation
value:
\begin{equation}
\min_G \max_D \; \E_{y\sim\rho_y}\left[\;\log D(y)\;\right] + \E_{z\sim p_z}\left[\;\log(1-D(G(z)))\;\right],
\label{eq:gan}
\end{equation}
by alternately maximizing the classification log-likelihood of $D$ and then
optimizing $G$ to deceive $D$ by minimizing the classification log-likelihood
of $1-D(G(z))$.
By this process, it is possible to
train the generative model to create realistic images from random
inputs~\cite{RadfordICLR2016}.  In practice, however, this min-max optimization
is unstable and hard to tune so that desired results can be obtained, which led
to some follow-up work~\cite{RadfordICLR2016,SalimansNIPS2016}.  This model is
also limited in that it can only handle relatively low, fixed resolutions.

This generative model was later extended to the Conditional Generative
Adversarial Networks (CGAN)~\cite{MirzaNIPSW2014}, which models $G:(x,z)
\mapsto y$ that generates $y$ conditioned on some input $x$.  Here, the
discriminator model $D:(x,y)\mapsto D(x,y)\in \mathbb{R}$ also takes $x$ as an
additional input to evaluate the \emph{conditional} probability given $x$.
Thus, $G$ and $D$ are optimized with the objective:
\begin{align}
\min_G \max_D \; &\E_{(x,y^*)\sim\rho_{x,y}}\left[\;\log D(x,y^*)\;\right]\nonumber\\
& + \E_{(x,y^*)\sim\rho_{x,y}, z\sim p_z}\left[\;\log(1-D(x,G(x,z)))\;\right].
\label{eq:cgan}
\end{align}
Note that the first expectation value is the average over supervised training
data $\rho_{x,y}$ consisting of input-output pairs. Because of this, the CGAN
framework is only applicable to supervised data.

The Conditional GAN can be used for the same purpose as ours, namely
\emph{prediction}, rather than sample generation.  For that, we just omit the
random input $z$, so that $S: x\mapsto y=S(x)$ is now a deterministic
prediction given the input $x$.  In this case, we thus optimize:
\begin{equation}
\min_S \max_D \; \E_{(x,y^*)\sim\rho_{x,y}}\left[\;\log D(x,y^*) + \log(1-D(x,S(x)))\;\right].
\label{eq:cganpred}
\end{equation}

\subsection{Adversarial Augmentation Framework}
In our view, the CGAN framework has one large limitation when used for the
structured prediction problem.  As we mentioned above, the CGAN objective
function \eqref{eq:cganpred} can only be trained with supervised training set,
because the crucial discriminator model $D$ is conditioned on the input $x$.
When data is hard to obtain, this becomes a significant hindrance to
performance.

\subsubsection{Supervised Adversarial Training}
To train the prediction model $S: x\mapsto y$ jointly with the discriminator model $D:y \mapsto D(y)\in \mathbb{R}$
that is \emph{not} conditioned on the input $x$, here we assume that the model $S$ has an associated
supervised training loss $L(S(x),y^*)$, where $y^*$ is the ground truth output corresponding to the input $x$.
We define the \emph{supervised adversarial training} as optimizing:
\begin{align}
\min_S \max_D \; & \E_{(x,y^*)\sim\rho_{x,y}}\left[\; \alpha \log D(y^*) \right. \nonumber \\
                 & \left.+ \alpha\log(1-D(S(x))) + L(S(x),y^*) \;\right] \quad,
\label{eq:discadvtrain}
\end{align}
where $\alpha$ is a weighting hyper-parameter and the expectation value is over a supervised training set
$\rho_{x,y}$ of input-output pairs.
This can be seen as a weighted combination of the loss $L(S(x),y^*)$ and a GAN-like adversarial loss,
trained on pairs of supervised samples.
The difference from GAN is that here we have supervised data, while GAN does not.
The difference from CGAN is that the coupling between the input $x$ and the ground truth output
$y^*$ is through the conditional discriminator $D$ in the case of CGAN (Eq.~\eqref{eq:cganpred}),
while in the case of the supervised adversarial training (Eq.~\eqref{eq:discadvtrain}),
 they are coupled directly through the supervised training loss $L(S(x),y^*)$.

The training consists of jointly maximizing the output of the discriminator
network $D$ and minimizing the loss of the model with structured output by $S$.
For each training iteration, we alternately optimize $D$ and $S$ until convergence.
The hyper-parameter $\alpha$ controls the influence of the adversarial training on the network and is
critical for training. Setting $\alpha$ too low gives no advantage 
over training with the supervised training loss, while setting it too high
causes the results to lose coherency with the inputs, \ie, the
network generates realistic outputs, however they do not correlate with the inputs.

\subsubsection{Unsupervised Data Augmentation}
Our objective function above is motivated by the desire for unsupervised training.
In most structured prediction problems, creating supervised training data by annotating the inputs is a time-consuming task.
However, in many cases, it is much easier to procure non-matching input and output data;
so it is desirable to be able to somehow use them to train our prediction model.
Note that in our objective function \eqref{eq:discadvtrain}, the first term inside the expectation value only needs $y$,
whereas the second only takes $x$.
This suggests that we can train using these terms separately with unsupervised data.
It turns out that we can indeed use the supervised adversarial objective function to also incorporate the unsupervised data
into the training, by separating the first two terms in the expectation value over supervised data in \eqref{eq:discadvtrain}
into new expectation values over unsupervised data. 

Suppose that we have large amounts of both input data $\rho_x$ and output data
$\rho_y$, in addition to a dataset $\rho_{x,y}$ of fully-annotated pairs. We modify
the optimization function to:
\begin{align}
\min_S \max_D \; &\E_{(x,y^*)\sim\rho_{x,y}}\left[\; L(S(x),y^*) + \alpha \log D(y^*) \right.\nonumber \\
                 &\qquad \left. + \alpha \log(1-D(S(x))) \;\right] \nonumber \\
                 &+ \beta \E_{y\sim\rho_y} \left[\; \log D(y) \;\right] \nonumber \\
                 &+ \beta \E_{x\sim\rho_x}\left[\; \log(1-D(S(x))) \;\right] \quad,
\label{eq:advaug}
\end{align}
where $\beta$ is a weighting hyper-parameter for the unsupervised data term.

\begin{table}[t]
   \begin{tabular}{rccc}
      \toprule
                 & Kernel & Activation & \\
      Layer Type &  Size  &  Function  & Output \\
      \midrule
      input         &      -     &  -  & $\phantom{00}1\times384\times384$ \\
      convolutional & $5\times5$ & ReLU & $\phantom{0}16\times192\times192$ \\
      convolutional & $3\times3$ & ReLU & $\phantom{0}32\times\phantom{0}96\times\phantom{0}96$ \\
      convolutional & $3\times3$ & ReLU & $\phantom{0}64\times\phantom{0}48\times\phantom{0}48$ \\
      convolutional & $3\times3$ & ReLU & $128\times\phantom{0}24\times\phantom{0}24$ \\
      convolutional & $3\times3$ & ReLU & $256\times\phantom{0}12\times\phantom{0}12$ \\
      dropout (50\%) &     -     &  -  & $512\times\phantom{00}6\times\phantom{00}6$ \\
      convolutional & $3\times3$ & ReLU &$512\times\phantom{00}6\times\phantom{00}6$ \\
      dropout (50\%) &     -     &  -  & $512\times\phantom{00}6\times\phantom{00}6$ \\
      fully connected &   -  & Sigmoid & $1$ \\
      \bottomrule
   \end{tabular}
   \vspace{-2mm}
   \caption{Architecture of the discriminator network. All convolutional layers
   use padding to avoid a decrease in output size and a stride of $2$ to half
   the resolution of the output.}
   \label{tbl:architecture}
\end{table}

Optimization is done on both $S$ and $D$ jointly using supervised data $\rho_{x,y}$
and unsupervised data $\rho_x$ and $\rho_y$. If learning from only $\rho_x$ and
$\rho_y$, the model $S$ will not necessarily learn the mapping $x \mapsto y$,
but only to generate realistic outputs $y$, which is not the objective of structured
prediction problems. Thus, using the supervised dataset $\rho_{x,y}$ is still
critical for training, as well as the model loss $L(S(x),y^*)$. The
supervised data can be seen as an anchor that forces the model to generate
outputs coherent with the inputs, while the unsupervised data is used to
encourage the model to generate realistic outputs for a wider variety of
inputs. See Fig.~\ref{fig:overview2} for a visualization of the approach. 
As we note above, it is not possible to train CGAN models \eqref{eq:cganpred}
using unsupervised data, as the discriminator network $D$ requires both the
input $x$ and output $y$ of the model as input.

One interesting benefit of being able to use unsupervised data is that, when after
training we use the prediction network to actually predict, we can use the input
to train the network on the fly to improve the results of the prediction.
This improves the prediction results by encouraging the prediction network $S$ to 
deceive the discriminator network $D$ and thus have more realistic outputs.
This does, however, incur a heavy overhead as it requires using the entire training
framework and optimizing the network.

\newcommand{\nfig}[1]{  \begin{adjustbox}{valign=t} \includegraphics[height=4.7cm]{figs/results_sketch/#1} \end{adjustbox} }
\newcommand{\nfigt}[1]{  \begin{adjustbox}{valign=t} \includegraphics[height=4.7cm]{figs/supplemental/#1} \end{adjustbox} }
\newcommand{\nfigp}[1]{ \begin{adjustbox}{valign=t} \includegraphics[height=4.7cm,clip,trim=40 0 0 0]{figs/results_sketch/#1} \end{adjustbox} }
\newcommand{\nfigs}[1]{ \includegraphics[height=2.35cm]{figs/results_sketch/#1} }
\begin{figure*}[th!]
\begin{center}
\setlength{\tabcolsep}{1pt}
\begin{tabular}{cccccc}
\begin{adjustbox}{valign=t} \raisebox{-31mm}{ \rotatebox{90}{Input Image} } \end{adjustbox} &
   \nfigt{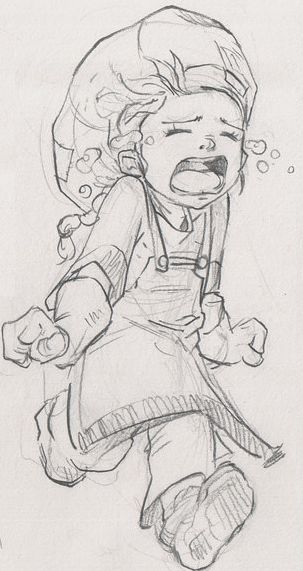} &
   \nfigt{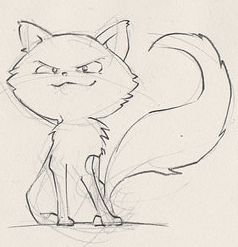} &
   \nfig{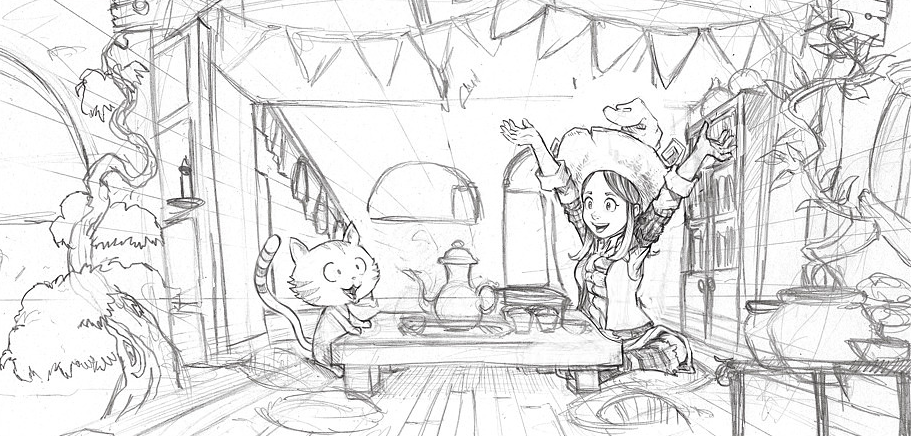} \\   \vspace{2mm}
\begin{adjustbox}{valign=t} \raisebox{-36mm}{ \rotatebox{90}{[Favreau et al. 2016]} } \end{adjustbox} &
   \nfigt{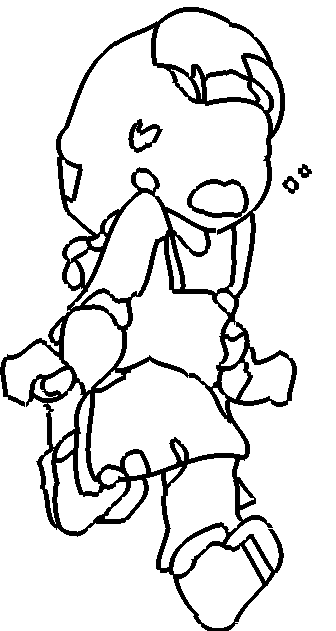} &
   \nfigt{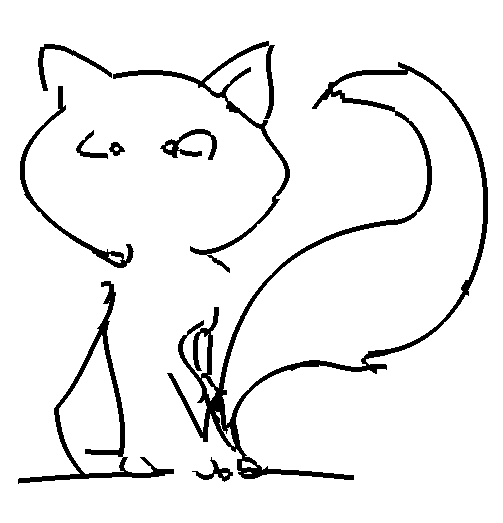} &
   \nfig{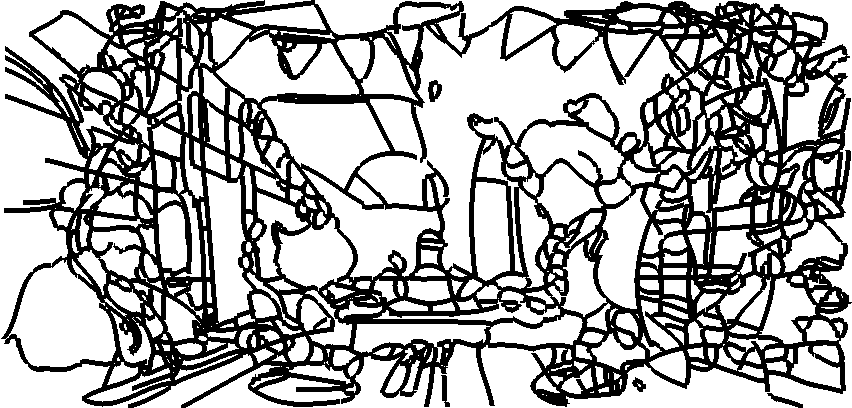} \\ \vspace{2mm}

\begin{adjustbox}{valign=t} \raisebox{-42mm}{ \rotatebox{90}{LtS [Simo-Serra et al. 2016]} } \end{adjustbox} &
   \nfigt{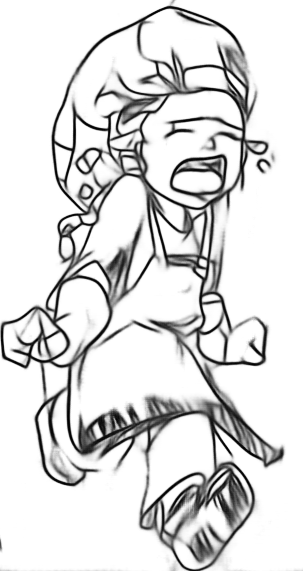} &
   \nfigt{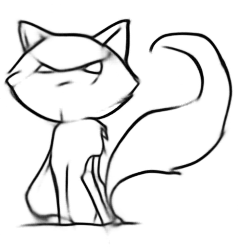} &
   \nfig{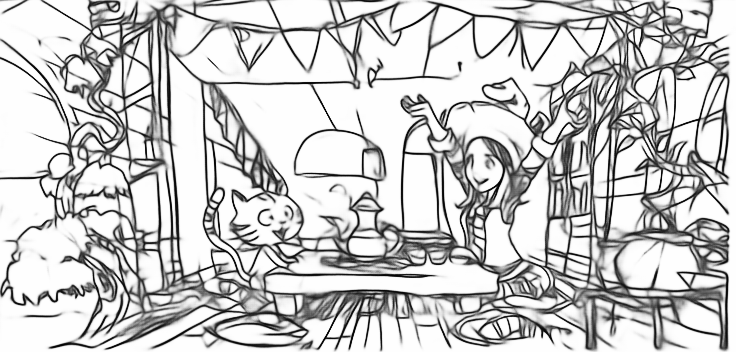} \\ \vspace{2mm}
\begin{adjustbox}{valign=t} \raisebox{-26mm}{ \rotatebox{90}{Ours} } \end{adjustbox} &
   \nfigt{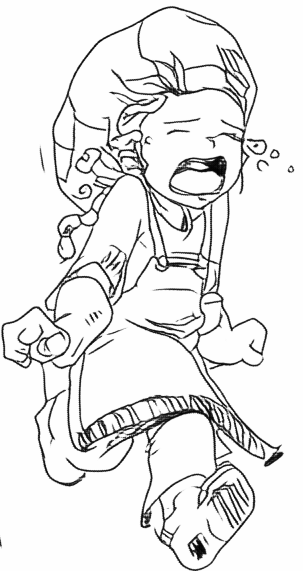} &
   \nfigt{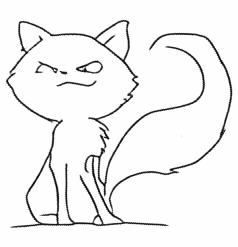} &
   \nfig{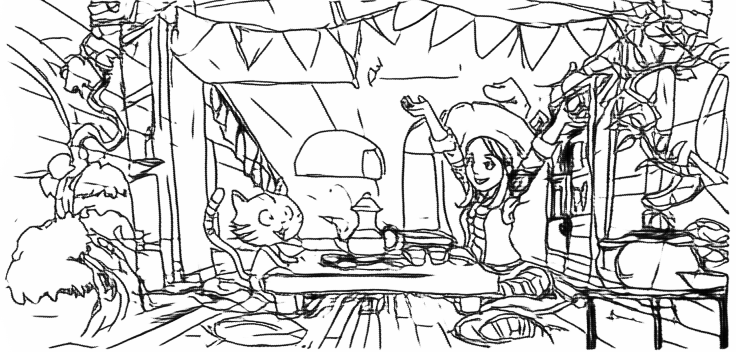} \\
\end{tabular}
\setlength{\tabcolsep}{6pt}
\end{center}
\vspace{-4mm}
\caption{Comparison against the state of the art methods of [Favreau et al. 2016] and
LtS [Simo-Serra et al. 2016]. For the approach of [Favreau et al. 2016], we had to
additionally preprocess the image with a tone curve and tune the default
parameters in order to obtain the shown results. Without this manual tweaking,
recognizable outputs were not obtained. For both LtS and our approach, we did
not preprocess the image and here we just visualize the network output. While [Favreau et
al. 2016] manages to capture the global structure somewhat, many different
parts of the image are missing due to the complexity of the scene. LtS fails to
simplify most regions in the scene and outputs blurry areas for low-confident
regions, which are dominant in these images. Our approach on the other hand is
able to simplify all images, both preserving detail and obtaining crisp and
clean outputs.
The images are copyrighted by \href{http://www.davidrevoy.com}{David Revoy
www.davidrevoy.com} and licensed under CC-by 4.0.
\vspace{-5mm}
}
\label{fig:stoa}
\end{figure*}

\section{Mastering Sketching}

We focus on applying our approach to the sketch simplification problem and its inverse.
Sketch simplification consists of processing rough
sketches such as those drawn by pencil into clean images that are amenable to
vectorization. Our approach is also the first approach that can handle
the inverse problem, that is, converting clean sketches into pencil drawings.

\subsection{Simplification Network}

In order to simplify rough sketches, we rely on the same model
architecture as \cite{SimoSerraSIGGRAPH2016}. The model consists of a 23-layer
fully-convolutional network that has three main building blocks:
down-convolutions, $3\times3$ convolutions with a stride of 2 to half the
resolution; flat-convolutions, $3\times3$ convolutions with a stride of 1 that
maintain the resolution; and up-convolutions, $4\times4$ convolutions with a
stride of $\nicefrac{1}{2}$ to double the resolution. In all cases, $1\times1$
padding to compensate the reduction in size caused by the convolution kernel
as well as the ReLU activation functions are employed. The general structure of the
network follows an hourglass shape, that is, the first seven layers contain three
down-convolution layers to decrease the resolution to one eighth. Afterwards, seven
flat-convolutions are used to further process the image, and finally, the last
nine layers contain three up-convolution layers to restore the resolution to that
of the input size. There are two exceptional layers: the first layer is a
down-convolution layer with a $5\times5$ kernel and $2\times2$ padding, and the
last layer uses a sigmoid activation function to output a greyscale image where
all values are in the $[0,1]$ range. In contrast with
\cite{SimoSerraSIGGRAPH2016}, which used a weighted MSE loss, we use the MSE
loss as the model loss
\begin{equation}
L(S(x),y^*)=\left\| \, S(x)-y^* \, \right\|^2 \quad,
\end{equation}
\noindent where $\|\cdot\|$ is the Euclidean norm.

Note that the MSE loss itself is not a structured prediction loss, \ie, it is oblivious of 
any structure between the component pixels. 
For each output pixel, neighboring output pixels have no effect. However, by additionally
using the supervised adversarial loss as done in Eq.~\eqref{eq:advaug}, the
resulting optimization does take into account the structure of the output.

\subsection{Discriminator Network}

The objective of the auxiliary discriminator network is not that of high
performance, but to help train the simplification network. If the discriminator
network becomes too strong with respect to the simplification network, the
gradients used for training generated by the discriminator network tend to
vanish, causing the optimization to fail to converge.  To avoid this issue, the
network is kept ``shallow'', uses large amounts of pooling, and employs
dropout~\cite{SrivastavaJMLR2014}. This also allows reducing the overall memory
usage and computation time, speeding up the training itself.

We base our discriminator network on a small CNN with seven layers, the last
being fully connected. Similarly to the simplification network, the first layer
uses a $5\times5$ convolution and all subsequent convolutional layers use
$3\times3$ convolutions. The first convolutional layer has 16 filters and all
subsequent convolutional layers double the number of filters. We
additionally incorporate 50\% dropout~\cite{SrivastavaJMLR2014} layers after
the last two convolutional layers. All fully-connected layers use Rectified
Linear Units (ReLU) except the final layer, which uses the sigmoid activation
function to have a single output that corresponds to the probability that the input
came from the real data $\rho_y$ instead of the output of $S$. An overview of
the architecture can be seen in Table~\ref{tbl:architecture}.

\subsection{Training}

Adversarial networks are notoriously hard to train; and
this has led to a series of heuristics for training. In particular, for
Generative Adversarial Networks (GAN), the balance between the learning of the
discriminative and generative components is critical, \ie, if the
discriminative component is too strong, the generative component is unable to
learn and vice versa. Unlike GAN, which has to rely entirely on the adversarial
network for learning, we also have supervised data $\rho_{x,y}$, which partially
simplifies the training.

Training of both networks is done with
backpropagation~\cite{RumelhartNATURE1986}. For a more fine-grained control of
the training, we balance the supervised training loss $L(S(x),y^*)$ and the
adversarial loss such that the gradients are roughly the same order of magnitude.

An alternate training scheme is used for both networks: in each iteration, we
first update the discriminator network with a mini-batch, and then proceed to
update the generative network using the same mini-batch.

During the training, we use batch normalization layers~\cite{IoffeICML2015} after
all convolutional layers for both the simplification network and the
discriminator network, which are then folded into the preceding convolutional
layers during the testing. Optimization is done using the ADADELTA
algorithm~\cite{ZeilerARXIV2012}, which abolishes the need for tuning hyper-parameters such
as the learning rate, adaptively setting a different learning rate for all the weights in the network.

We follow a similar data augmentation approach as~\cite{SimoSerraSIGGRAPH2016},
namely training with eight additional levels of downsampling: \nicefrac{7}{6}, \nicefrac{8}{6},
\nicefrac{9}{6}, \nicefrac{10}{6}, \nicefrac{11}{6}, \nicefrac{12}{6},
\nicefrac{13}{6}, and \nicefrac{14}{6}, while using the constant-size $384\times384$ image patch crops.
All training output images are thresholded so
that pixel values below 0.9 are set to 0 (pixels are in the $[0,1]$ range). All
the images are randomly rotated and flipped during the training.
Furthermore, we sample the image patches with more probability from larger images,
such that patches from a $1024\times1024$ image will be four times more likely
to appear than patches from a $512\times512$ image. We subtract the mean of the
input images of the supervised dataset from all images. Finally, with a
probability of 10\%, the ground truth images are used as both input and output
with the supervised loss, to teach the model that sketches that are already
simplified should not be modified.

\begin{figure}[t]
\begin{center}
   \begin{subfigure}[b]{0.37\linewidth}
      \centering
      \includegraphics[height=3.4cm]{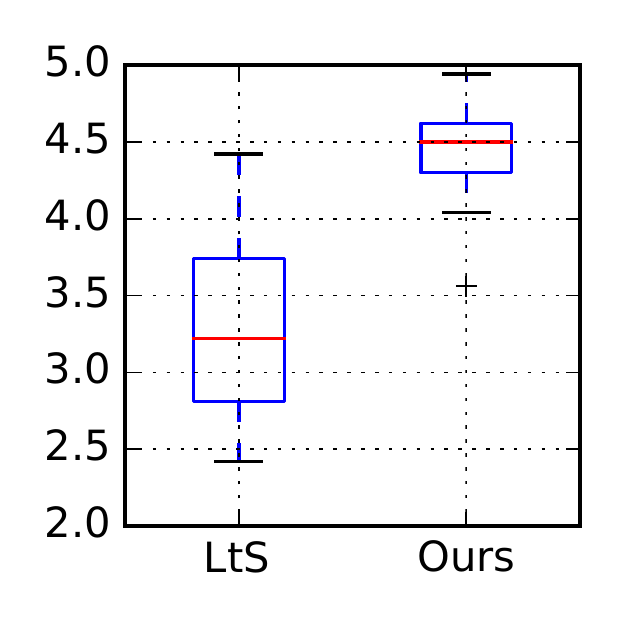}
      \vspace{-7mm}
      \caption{Absolute rating.}
   \end{subfigure}
   \begin{subtable}[b]{0.6\linewidth}
      \centering
      \begin{tabular}{rcc}
         \toprule
         & LtS & Ours \\
         \cmidrule{2-3}
         absolute & 3.28 & 4.43 \\
         \cmidrule{2-3}
         vs LtS   &    -   & 92.3\% \\
         vs Ours  &  7.7\% &    -   \\
         \bottomrule
      \end{tabular}
      \vspace{3mm}
      \caption{Mean results for all users.}
   \end{subtable}
\end{center}
\vspace{-2mm}
\caption{Results of the user study in which we evaluate the state of the
art of LtS [Simo-Serra et al. 2016] 
and our approach on both absolute (1-5 scale)
and relative (which result is better?) scales.}
\label{fig:results}
\end{figure}

\renewcommand{\nfig}[1]{ \includegraphics[width=0.32\linewidth]{figs/supplemental/#1} }
\begin{figure}[t]
\begin{center}
\setlength{\tabcolsep}{0pt}
\begin{tabular}{ccc}
   {\footnotesize Input} & {\footnotesize LtS} & {\footnotesize Ours} \\
   \nfig{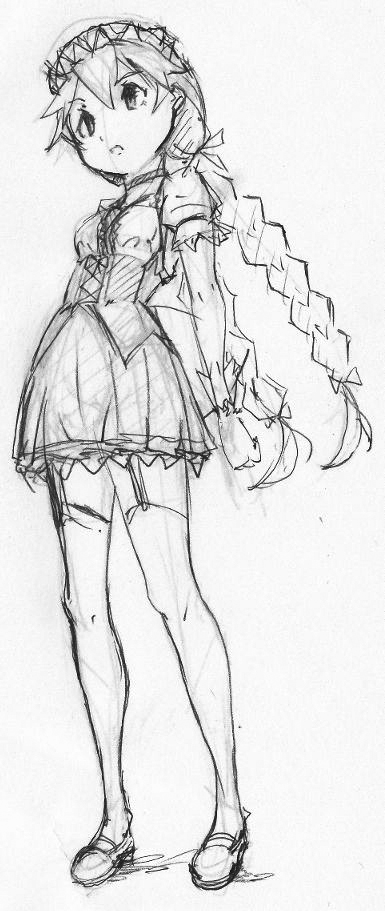} &
   \nfig{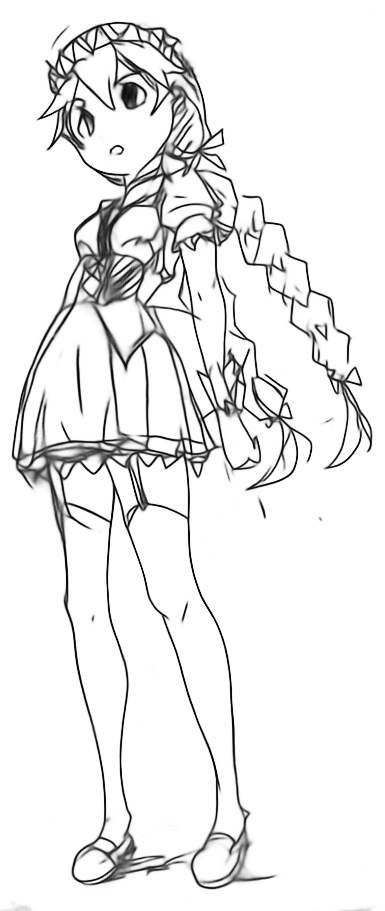} &
   \nfig{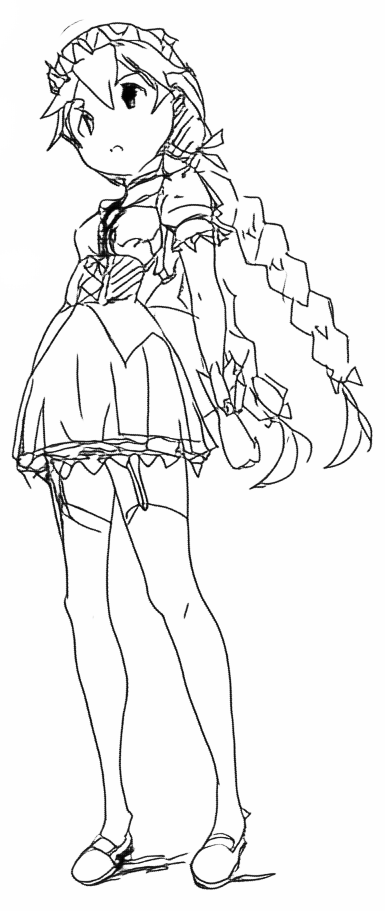} \\   
\end{tabular}
\setlength{\tabcolsep}{6pt}
\end{center}
\vspace{-3mm}
\caption{Example results included in the user study. The drawing is copyrighted by flickr user ``Yama Q'' and licensed under CC-by-nc 4.0}
\label{fig:results_ex}
\end{figure}

\renewcommand{\nfig}[1]{ \includegraphics[width=0.24\linewidth]{figs/rough/#1} }
\begin{figure*}[th]
\begin{center}
\setlength{\tabcolsep}{0pt}
\begin{tabular}{cccc}
{\normalsize Input} & {\normalsize MSE Loss} & {\normalsize Adv. Aug. (Artist 1)} & {\normalsize Adv. Aug. (Artist 2)} \\
\nfig{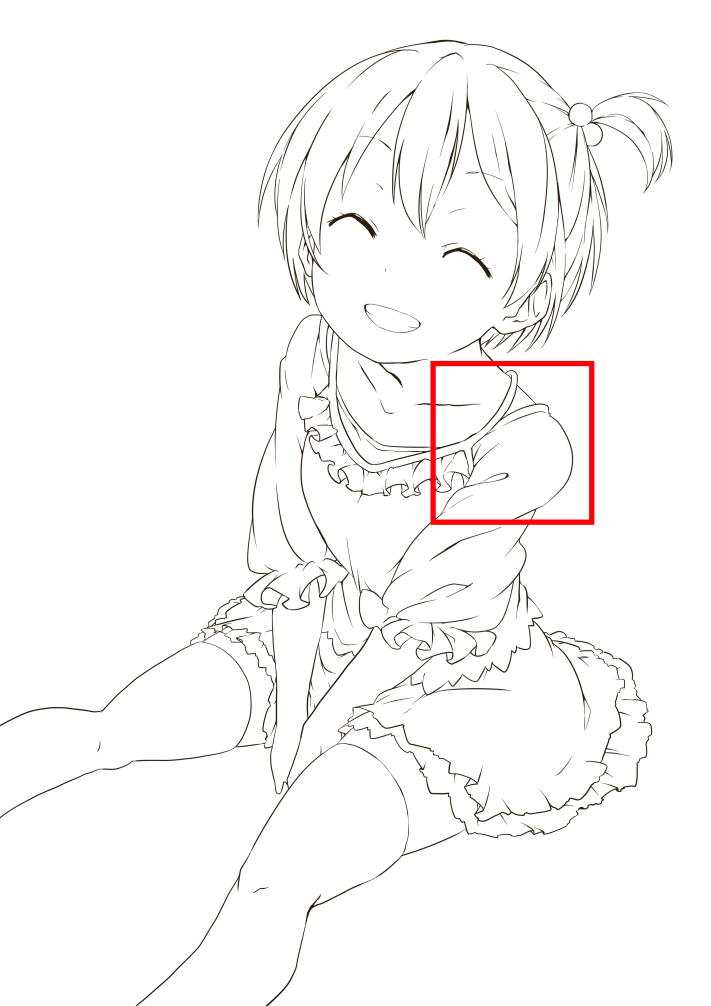} &
\nfig{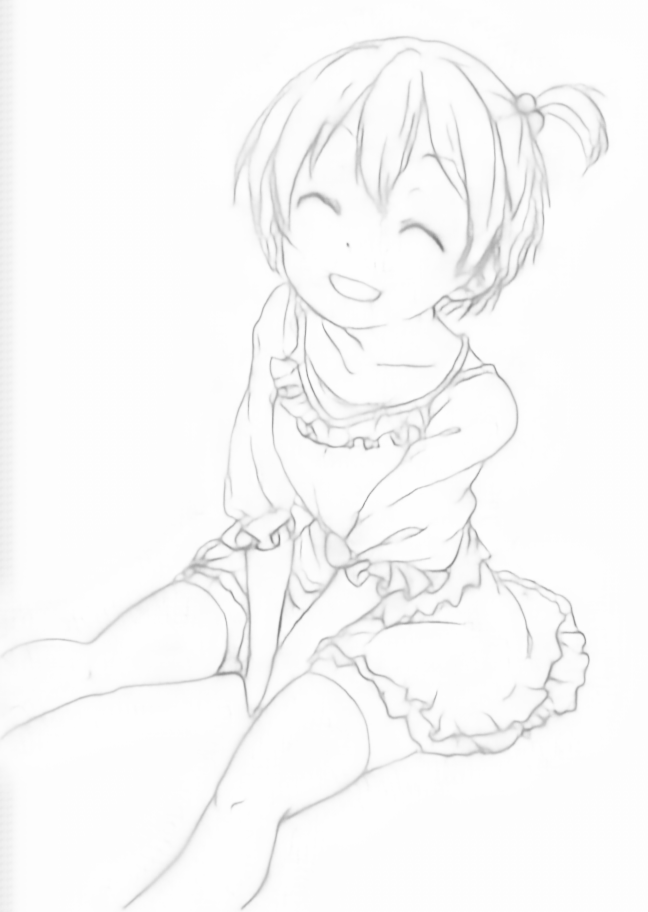} &
\nfig{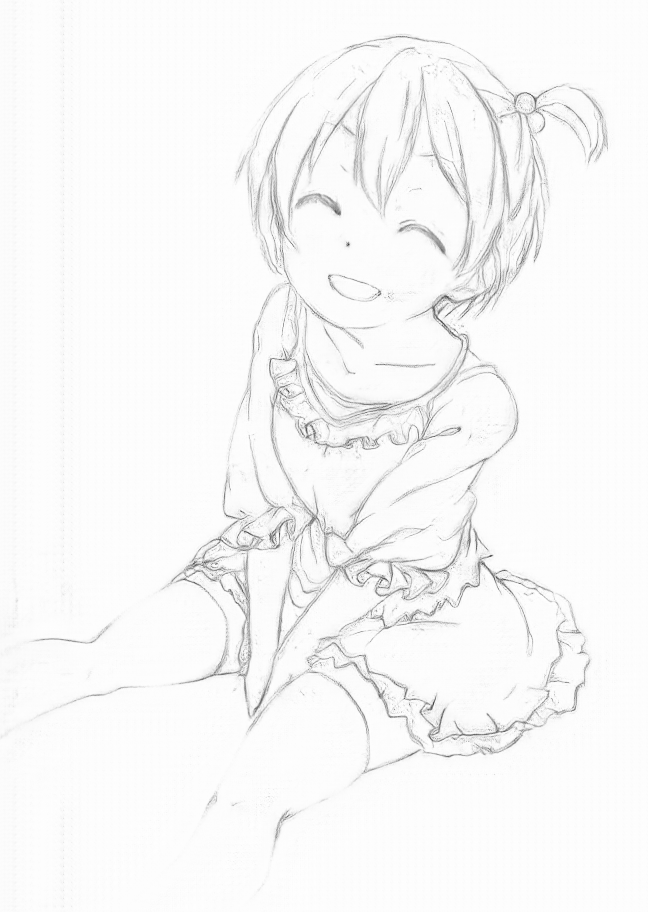} &
\nfig{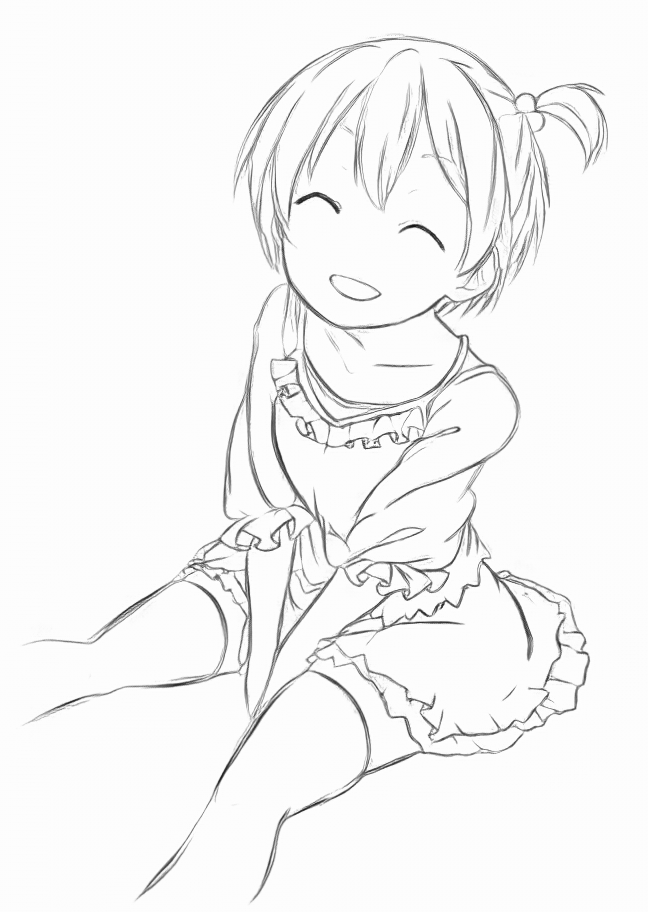} \\
\nfig{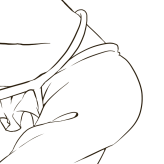} &
\nfig{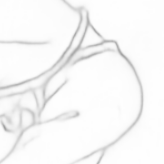} &
\nfig{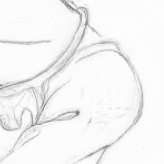} &
\nfig{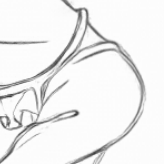} \\
\end{tabular}
\setlength{\tabcolsep}{6pt}
\end{center}
\vspace{-3mm}
\caption{Examples of pencil drawing generation with our training framework. We compare
three models: one trained with the standard MSE loss, and two models trained
with adversarial augmentation using data from two different artists. In the first column, we
show the input to all three models, followed by the outputs of each
model. The first row shows the entire image, while the bottom row shows the
area highlighted in red in the input image zoomed. We can see that the MSE loss
only succeeds in blurring the input image, while the two models trained with
adversarial augmentation are able to show realistic pencil drawings. We also
show how training on data from different artists gives significantly different
results. Artist 1 tends to add lots of smudge marks even far away from the
lines, while artist 2 uses many overlapping lines to give the shape and form to
the drawing.}
\label{fig:roughsketch}
\end{figure*}

\renewcommand{\nfig}[1]{ \includegraphics[width=0.24\linewidth]{figs/rough2/#1} }
\begin{figure*}[th]
\includegraphics[width=\linewidth]{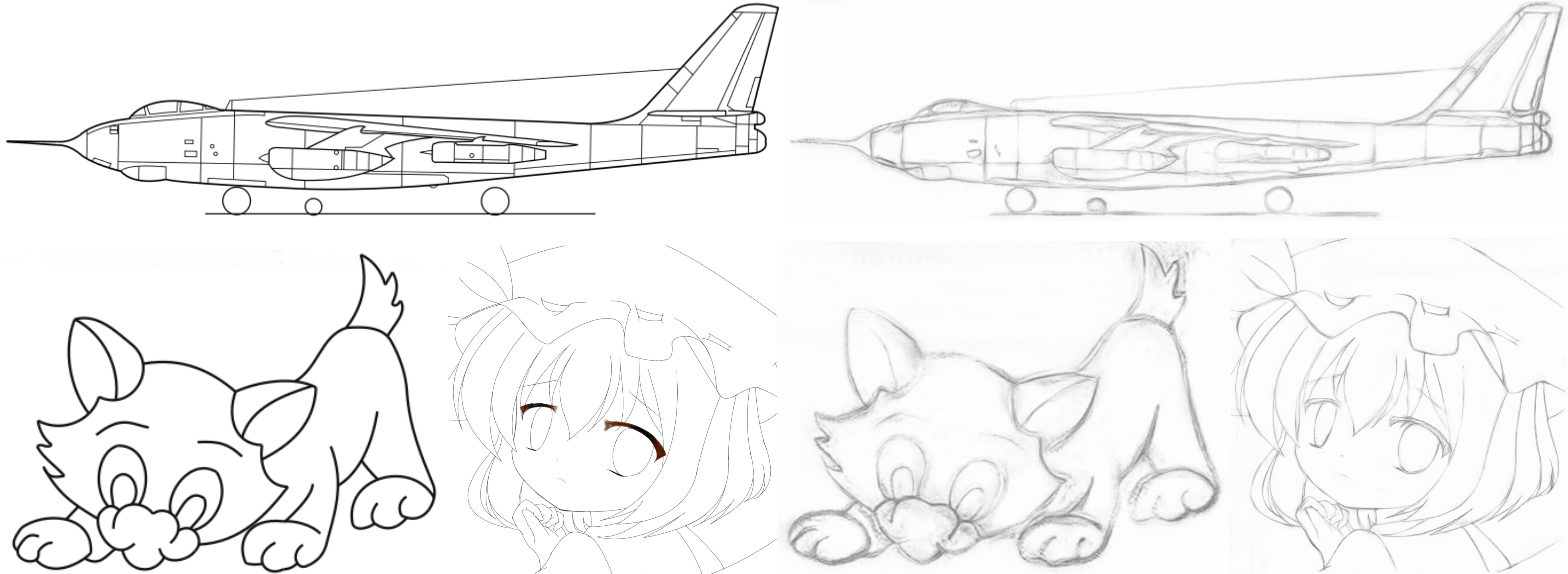}
\vspace{-6mm}
\caption{More examples of pencil drawing generation. The line drawings on the
left are automatically converted to the pencil drawings on the right.}
\label{fig:moreroughsketch}
\end{figure*}

\newcommand{\mfig}[1]{\includegraphics[height=2.8cm]{figs/results_sketch/#1}}
\newcommand{\mfigs}[1]{\raisebox{0.3cm}{\includegraphics[height=2.2cm]{figs/results_sketch/#1}}}
\newcommand{\mfigb}[1]{\includegraphics[height=2.3cm]{figs/results_sketch/#1}}
\begin{figure*}[t]
\begin{center}
\setlength{\tabcolsep}{0pt}
\begin{tabular}{cccccc}
   \multicolumn{2}{c}{Input} & \multicolumn{2}{c}{Supervised-only} & \multicolumn{2}{c}{Ours} \\
   \mfigs{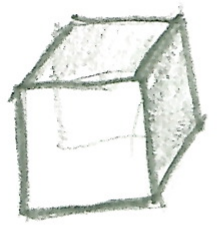} &
   \mfig{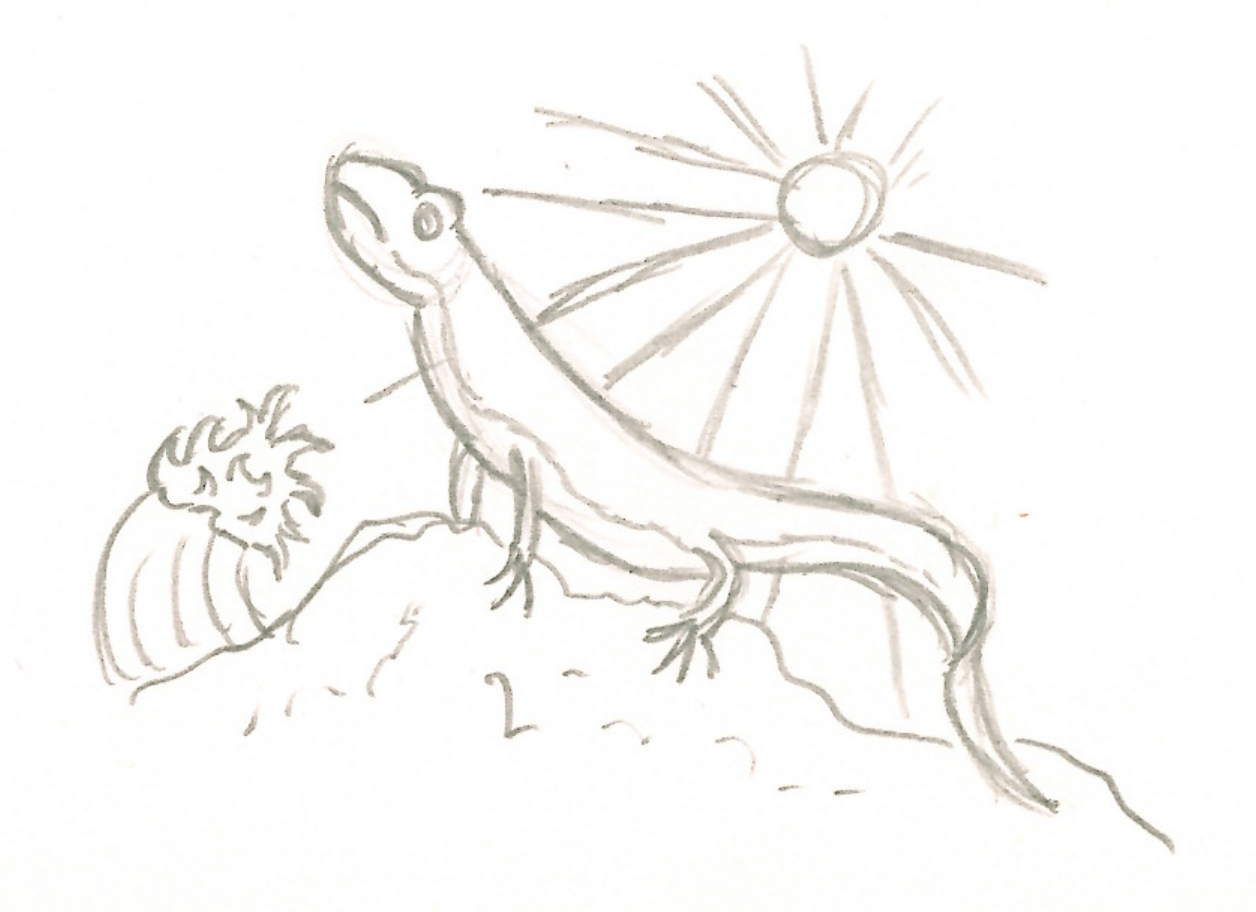} &
   \mfigs{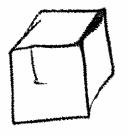} &
   \mfig{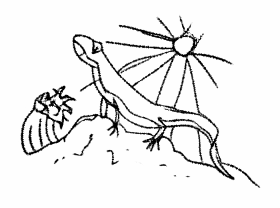} &
   \mfigs{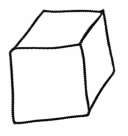} &
   \mfig{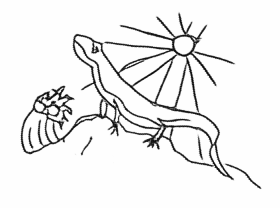} \\
   \multicolumn{2}{c}{\mfigb{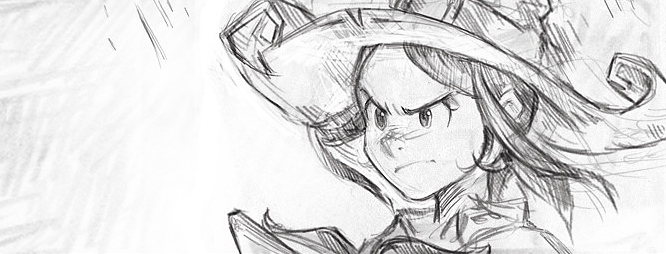}} &
   \multicolumn{2}{c}{\mfigb{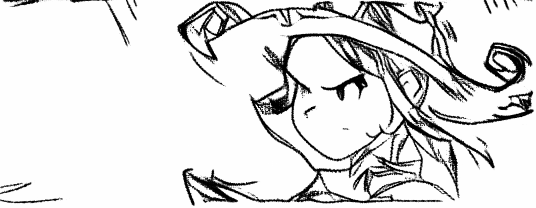}} &
   \multicolumn{2}{c}{\mfigb{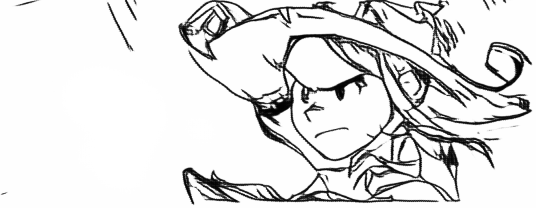}} \\
\end{tabular}
\setlength{\tabcolsep}{6pt}
\end{center}
\vspace{-3mm}
\caption{Visualization of the benefits of using additional unsupervised data
for training with our approach. For rough sketches fairly different from those
in the training data we can see a clear benefit when using additional
unsupervised data. Note that this data, in contrast with supervised data, is
simple to obtain. We note that other approaches such as CGAN are unable to use
unsupervised data in training.
The bottom image is copyrighted by
\href{http://www.davidrevoy.com}{David Revoy www.davidrevoy.com} and licensed under CC-by 4.0.
}
\label{fig:unsupervised}
\vspace{-3mm}
\end{figure*}

\section{Experiments}

We train our models using the supervised dataset
from~\cite{SimoSerraSIGGRAPH2016}, consisting of 68 pairs of rough sketches
with their corresponding clean sketches ($\rho_{x,y}$), in addition to 109
clean sketches ($\rho_y$) and 85 rough sketches ($\rho_x$) taken from Flickr and other sources.
Note that the additional training data $\rho_y$ and $\rho_x$ require no
additional annotations, unlike the original supervised dataset. Some examples
of the data used for training are shown in Fig.~\ref{fig:overview2}. We
set $\alpha=\beta=8\times10^{-5}$ and train for 150,000
iterations. Each batch consists of 16 pairs of image patches sampled from the
68 pairs in $\rho_{x,y}$, 16 image patches sampled from the 109 clean sketches in
$\rho_y$, and 16 image patches sampled from the 85 rough sketches in $\rho_x$. We
initialize the model weights for all models by training exclusively in a
supervised fashion on the supervised data $\rho_{x,y}$, and in particular use
the state-of-the-art model~\cite{SimoSerraSIGGRAPH2016}. We note this
pretraining is critical for learning and that without it training is greatly
slowed down and in the case of CGAN it does not converge.

\subsection{Comparison against the State of the Art}

We compare against the state of the art of \cite{FavreauSIGGRAPH2016} and
Learning to Sketch (LtS)~\cite{SimoSerraSIGGRAPH2016} in Fig.~\ref{fig:stoa}.
We found that the post-processing done in LtS requires per-image tuning for
images ``in the wild'' and opt to show directly the model outputs instead.
We can see that in general the approach of \cite{FavreauSIGGRAPH2016} fails to
conserve most fine details, while conserving unnecessary details. On the other
hand, LtS has low confidence on most fine details resulting in large blurry
sections. Our proposed approach is able to both conserve fine details and avoid
blurring.

\subsection{User Study}

We perform two user studies for a quantitative analysis on additional test data
that is not part of our unsupervised training set. For both studies, we process
50 images with both our approach and LtS. In the
first study, we randomly show the output of both approaches side-by-side to 15
users, asking them to choose the better result, while in the second study we show
both the input rough sketch and a simplified sketch and ask them to rate the
conversion on a scale of 1 to 5. We directly display the output of both networks; and the
order of the images shown is randomized for every user. Evaluation results with
an example of evaluated images are shown in Fig.~\ref{fig:results}.

In the absolute evaluation we can see that, while both approaches are scored
fairly high, our approach obtains 1.15 points above the state of the art on a
scale of 1 to 5. In the relative evaluation, our approach is preferred to the
state of the art 92.3\% of the time, with 7 of the 13 users preferring our
approach 100\% of the time, and the lowest scoring user preferring our approach
72\% of the time. This highlights the importance of using adversarial
augmentation to obtain more realistic sketch outputs, avoiding blurry or
ill-defined areas. From the example images, we can see that the LtS model
in general tends to blur complicated areas it is not able to fully
parse, while our approach always produces well-defined and crisp outputs.
Note that both network architectures are exactly the same: only the learning
process and thus the weight values change.  Additional qualitative examples are
shown in Fig.~\ref{fig:stoa}, where it can be seen that our approach outputs
sketch simplifications without blurring.

\subsection{Pencil Drawing Generation}

We also apply our proposed approach to the inverse problem of sketch
simplification, that is, pencil drawing generation. We swap
the input and output of the training data used for sketch simplification and
train new models. However, unlike sketch simplification, it turns out that
it is not possible to obtain realistic results without supervised
adversarial training: the output just becomes a blurred version of the input. Only
by introducing the adversarial augmentation framework is it possible
to learn to produce realistic pencil sketches. We train three models: one with
the MSE loss, and two with adversarial augmentation for different artists. MSE
loss and Artist 1 models are trained on 22 image pairs, while the Artist 2
model is trained on 80 image pairs. We do not augment the training data with
unsupervised examples, as we only have training pairs for both artists. Results
are shown in Fig.~\ref{fig:roughsketch}. We can see how the adversarial
augmentation is critical in obtaining realistic outputs and not just a blurred
version of the input. Furthermore, by training on different artists, we seem to
obtain models that capture each artists' personality and nuances. Additional
results are shown in Fig.~\ref{fig:moreroughsketch}.

\subsection{Generalizing with Unsupervised Data}

One of the main advantages of our approach is the ability to exploit
unsupervised data. This is very beneficial as acquiring matching pairs of rough
sketches and simplified sketches is very time consuming and laborious.
Furthermore, it is hard to obtain examples from many different illustrators to
teach the model to simplify a wide variety of styles. We train a model using
the supervised adversarial loss, \ie, without unsupervised data, by setting
$\beta=0$ and compare against our full model using unsupervised data in
Fig.~\ref{fig:unsupervised}. We can see a clear benefit in images fairly
different from those in the training data, indicating better generalization of
the model. In contrast to our approach, existing approaches are unable to
benefit from a mix of supervised and unsupervised data.

\subsection{Single-Image Optimization}

\newcommand{\sfig}[1]{ \includegraphics[height=4.3cm]{figs/singleOpt/#1} }
\begin{figure*}[t]
\begin{center}
\setlength{\tabcolsep}{1pt}
\begin{tabular}{cccccc}
{\normalsize Input} &
{\normalsize Output} &
{\normalsize Optimized} &
{\normalsize Input} &
{\normalsize Output} &
{\normalsize Optimized} \\
\sfig{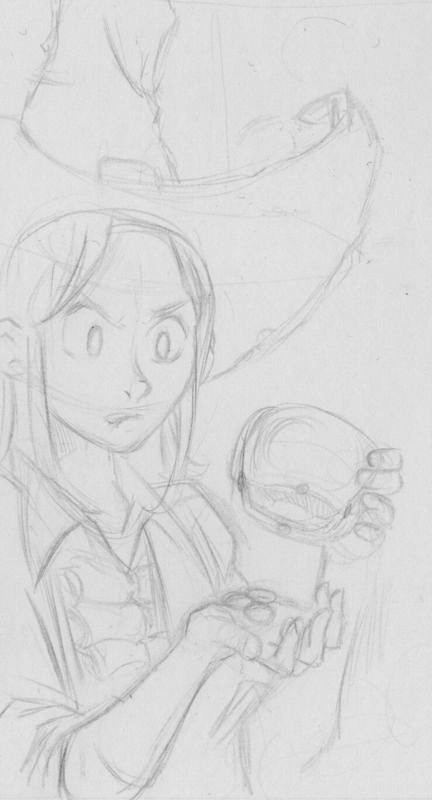} & 
\sfig{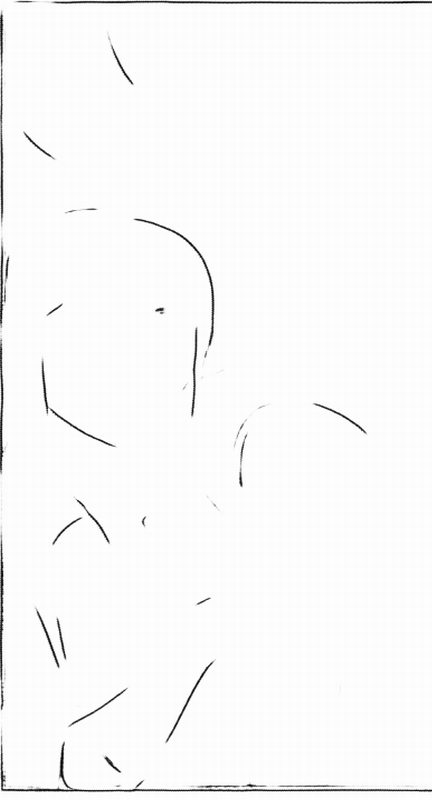} &
\sfig{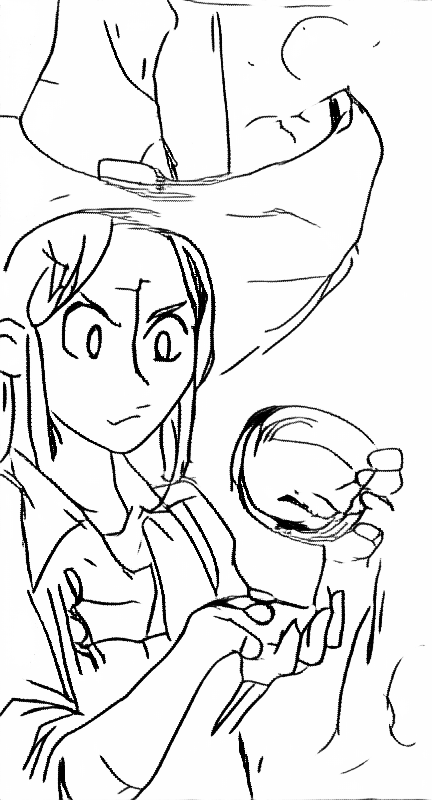} &
\sfig{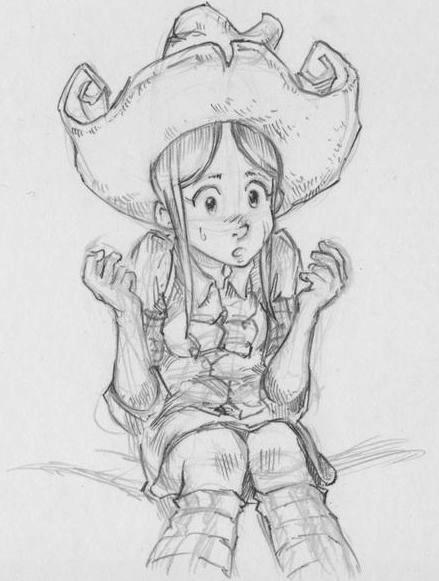} & 
\sfig{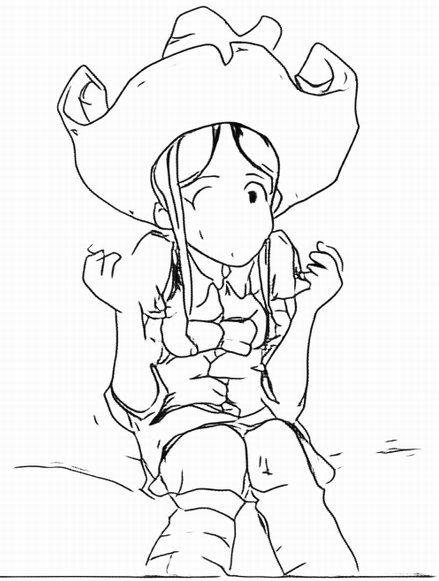} &
\sfig{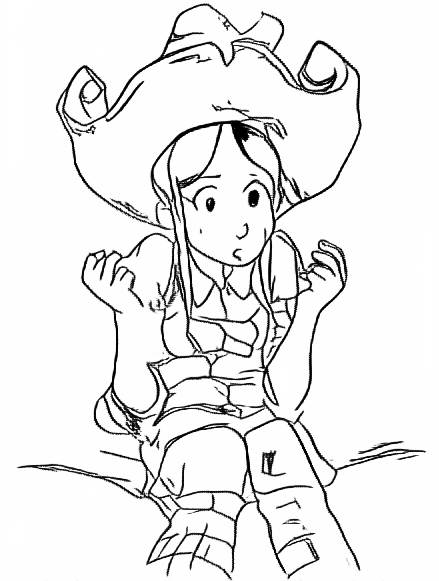} \\[-1mm]
\end{tabular}
\setlength{\tabcolsep}{6pt}
\end{center}
\vspace{-3mm}
\caption{Single image optimization. We show examples of images in which our
proposed model does not obtain very good results, and optimize our model for
these single images in an unsupervised fashion. This optimization process
allows adapting the model to new data without annotations.
The images are copyrighted by
\href{http://www.davidrevoy.com}{David Revoy www.davidrevoy.com} and licensed under CC-by 4.0.
}
\label{fig:singleopt}
\vspace{-2mm}
\end{figure*}

As another extension of our framework, we introduce the single-image
optimization. Since we are able to directly use unsupervised data, it seems
natural to use the test set with the adversarial augmentation framework to
optimize the model for the test data. Note that this is done in the test time
and does not involve any privileged information as the test set is used in a
fully unsupervised manner. We test this approach using a single additional
image and optimizing the network for this image. Optimization is done by using
the adversarial augmentation from Eq.~\eqref{eq:advaug} with $\alpha=0$,
$\rho_y \subset \rho_{x,y}$; with $\rho_x$ consisting of the single test image.
The other hyper-parameters are
set to the same values as used for sketch simplification. Results are shown in
Fig.~\ref{fig:singleopt}. We can see how optimizing results on a single test
image can provide a further increase in accuracy, particularly when considering
very hard images. In particular, in the left image, using the pretrained model
leads to a non-recognizable output, as there is very little contrast in the
input image. We do note, however, that this procedure leads to inference times
a few orders of magnitude slower than using a pretrained network.

\subsection{Comparison against Conditional GAN}

\renewcommand{\nfig}[2]{
   \includegraphics[width=0.32\linewidth]{figs/results_sketch/#1} &
   \includegraphics[width=0.32\linewidth]{figs/results_sketch/#1_#2_real_cgan_iter0051000_f} &
   \includegraphics[width=0.32\linewidth]{figs/results_sketch/#1_#2_sig2017_f} }
\newcommand{\nfige}[2]{
   \includegraphics[width=0.32\linewidth]{figs/results_extra/#1} &
   \includegraphics[width=0.32\linewidth]{figs/results_extra/#1_#2_real_cgan_iter0051000_f} &
   \includegraphics[width=0.32\linewidth]{figs/results_extra/#1_#2_sig2017_f} }
\newcommand{\nfigc}[1]{
   \includegraphics[width=0.32\linewidth]{figs/cgan/#1_rough} &
   \includegraphics[width=0.32\linewidth]{figs/cgan/#1_cgan} &
   \includegraphics[width=0.32\linewidth]{figs/cgan/#1_ours} }
\begin{figure}[t]
\begin{center}
\setlength{\tabcolsep}{0pt}
\begin{tabular}{ccc}
   {\normalsize Input} & {\normalsize CGAN} & {\normalsize Ours} \\[-1mm]
   \nfig{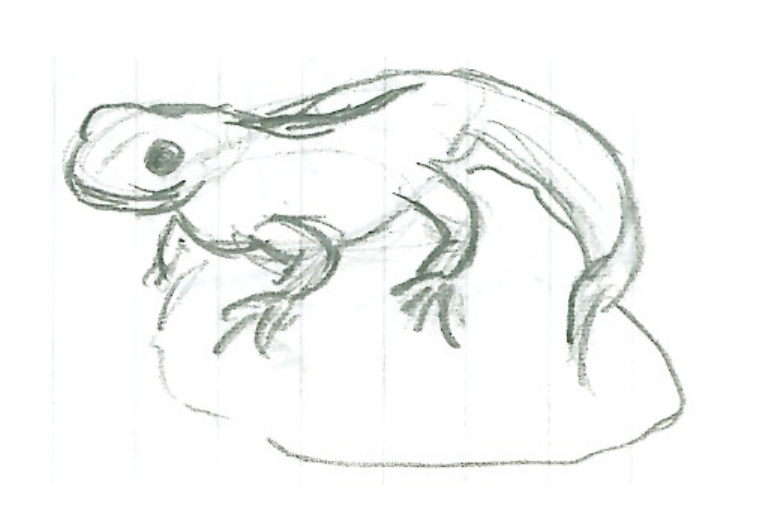}{s450} \\[-2mm]
   \nfig{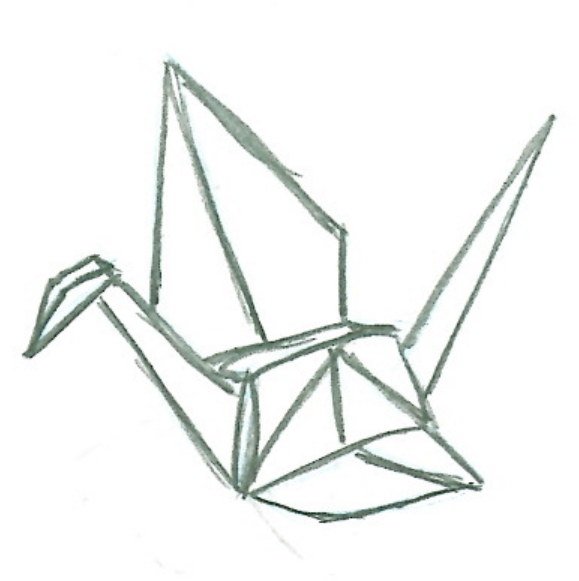}{s200} \\[-3mm]
   \nfigc{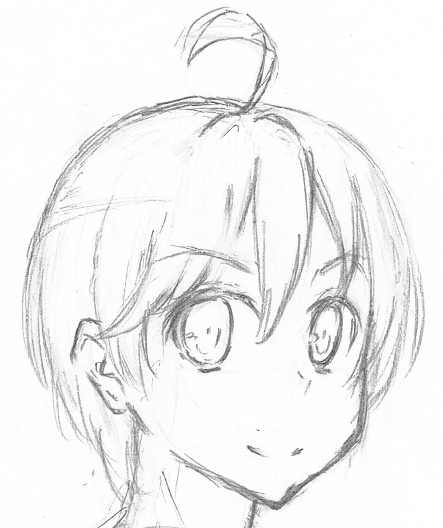} \\
   \nfigc{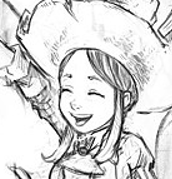} \\
\end{tabular}
\setlength{\tabcolsep}{6pt}
\end{center}
\vspace{-4mm}
\caption{Comparison of our approach against the Conditional GAN approach.
The bottom image is copyrighted by \href{http://www.davidrevoy.com}{David Revoy
www.davidrevoy.com} and licensed under CC-by 4.0.
}
\label{fig:comp_cgan}
\vspace{-2mm}
\end{figure}

We also perform a qualitative comparison against the recent Conditional GAN
(CGAN) approach as an alternative learning scheme. As in
the other comparisons, the CGAN is pretrained using the model of
\cite{SimoSerraSIGGRAPH2016}. The training data is the same as our model when
using only supervised data, the difference lies in the loss. The CGAN model
uses a loss based on Eq.~\eqref{eq:cganpred}, while the supervised model uses
Eq.~\eqref{eq:discadvtrain}. The discriminator network of the
CGAN model uses both the rough sketch $x$ and the simplified sketch $y$ as an
input, while in our approach $D$ only uses the simplified sketch $y$. We note that we
found the CGAN model to be much more unstable during training, several times
becoming completely unstable forcing us to redo the training. This is likely
caused by only using the GAN loss in contrast with our model that also uses the
MSE loss for training stability.

Results are shown in Fig.~\ref{fig:comp_cgan}. We can see that the CGAN
approach is able to produce non-blurry crisp lines thanks to the GAN loss,
however, it fails at simplifying the input image and adds additional artefacts.
This is likely caused by the GAN loss itself, as it is a very unstable loss
prone to artefacting. Our approach on the other hand uses a different loss that
also includes the MSE loss which adds stability and coherency to the output
images.

\subsection{Discussion}

\newcommand{\nfigu}[2]{
   \includegraphics[width=0.32\linewidth]{figs/unsup/esimo_#1} &
   \includegraphics[width=0.32\linewidth]{figs/unsup/esimo_#1_#2_cgan_iter0133000_f} &
   \includegraphics[width=0.32\linewidth]{figs/unsup/esimo_#1_#2_sig2017_f} }
\begin{figure}[t]
\begin{center}
\setlength{\tabcolsep}{0pt}
\begin{tabular}{ccc}
   {\normalsize Input} & {\normalsize Only Unsupervised} & {\normalsize Ours} \\
   \nfigu{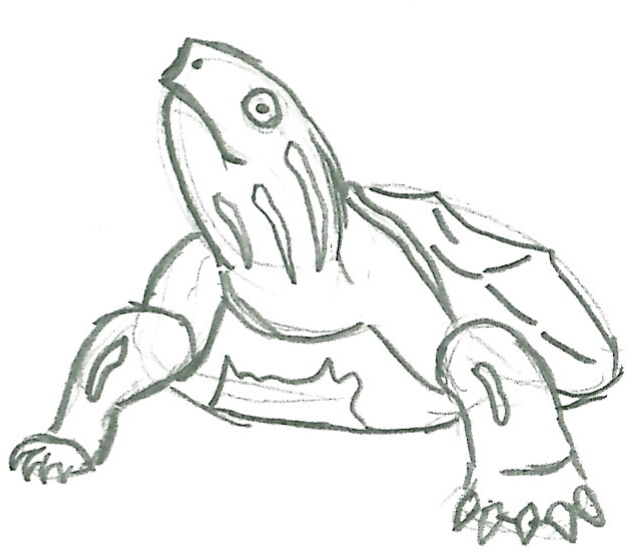}{s400} \\
   \nfigu{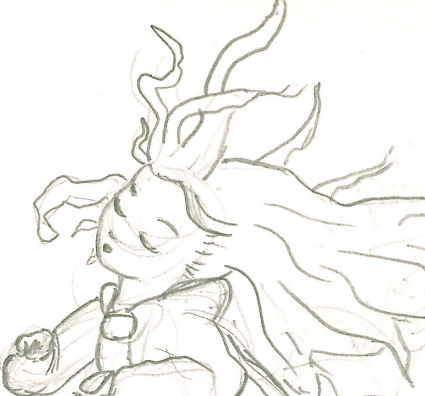}{s500} \\
\end{tabular}
\setlength{\tabcolsep}{6pt}
\end{center}
\vspace{-4mm}
\caption{Comparison of our approach with and without supervised data. With only
unsupervised data, the output loses its coherency with the input and ends up
looking like abstract line drawings.}
\label{fig:comp_unsup}
\vspace{-2mm}
\end{figure}

While our approach can make great use of unsupervised data, it still has
an important dependency on high quality supervised data, without which it would
not be possible to obtain good results. As an extreme case, we train a model
without supervised data and show results in Fig.~\ref{fig:comp_unsup}. Note
that this model uses the initial weights of the LtS model, without which it
would not be possible to train it. While the output images do look like line
drawings, they have lost any coherency with the input rough sketch.


\section{Conclusions}

We have presented the adversarial augmentation for structured prediction and
applied it to the sketch simplification task as well as its inverse problem, \ie,
pencil drawing generation. We have shown that by augmenting the standard model
loss with a supervised adversarial loss, it is possible to get much more
realistic structured outputs. Furthermore, the same framework allows for
unsupervised data augmentation, essential for structured prediction tasks in which
obtaining additional annotated training data is very costly. The proposed
approach also allows opening the door to tasks that are not possible with
standard losses such as generating pencil drawings from clean sketches and has
wide applicability to most structured prediction problems. As adversarial
augmentation only applies to the training, the resulting models have exactly the
same inference properties as the non-augmented versions. As a further extension
of the problem, we show that the framework can also be used to optimize for a
single input for situations in which accuracy is valued more than quick
computation. This can, for example, be used to personalize the model to
different artists using only unsupervised rough and clean training data from
each particular artist.

\section{Acknowledgements}
This work was partially supported by JST CREST and JST ACT-I Grant Number JPMJPR16UD.

\bibliographystyle{acmsiggraph}
\bibliography{references}

\begin{thebibliography}{\protect\citename{Goodfellow et~al\mbox{.} }2014}

\bibitem[\protect\citename{Bae et~al\mbox{.}
  }2008]{Bae:2008:IAS:1449715.1449740}
{\sc Bae, S.-H., Balakrishnan, R., and Singh, K.}
\newblock 2008.
\newblock Ilovesketch: As-natural-as-possible sketching system for creating 3d
  curve models.
\newblock In {\em ACM Symposium on User Interface Software and Technology},
  151--160.

\bibitem[\protect\citename{Berger et~al\mbox{.} }2013]{BergerSIGGRAPH2013}
{\sc Berger, I., Shamir, A., Mahler, M., Carter, E., and Hodgins, J.}
\newblock 2013.
\newblock Style and abstraction in portrait sketching.
\newblock {\em ACM Transactions on Graphics 32}, 4, 55.

\bibitem[\protect\citename{Chen et~al\mbox{.} }2013]{ChenCGF2013}
{\sc Chen, J., Guennebaud, G., Barla, P., and Granier, X.}
\newblock 2013.
\newblock Non-oriented mls gradient fields.
\newblock {\em Computer Graphics Forum 32}, 8, 98--109.

\bibitem[\protect\citename{Dong et~al\mbox{.} }2016]{DongPAMI2016}
{\sc Dong, C., Loy, C.~C., He, K., and Tang, X.}
\newblock 2016.
\newblock Image super-resolution using deep convolutional networks.
\newblock {\em IEEE Transactions on Pattern Analysis and Machine Intelligence
  38}, 2, 295--307.

\bibitem[\protect\citename{Favreau et~al\mbox{.} }2016]{FavreauSIGGRAPH2016}
{\sc Favreau, J.-D., Lafarge, F., and Bousseau, A.}
\newblock 2016.
\newblock Fidelity vs. simplicity: a global approach to line drawing
  vectorization.
\newblock {\em ACM Transactions on Graphics (Proceedings of SIGGRAPH) 35}, 4.

\bibitem[\protect\citename{Fi\v{s}er et~al\mbox{.} }2015]{FiserSketch2015}
{\sc Fi\v{s}er, J., Asente, P., and S\'{y}kora, D.}
\newblock 2015.
\newblock Shipshape: A drawing beautification assistant.
\newblock In {\em Workshop on Sketch-Based Interfaces and Modeling}, 49--57.

\bibitem[\protect\citename{Fukushima }1988]{FukushimaNN1988}
{\sc Fukushima, K.}
\newblock 1988.
\newblock Neocognitron: A hierarchical neural network capable of visual pattern
  recognition.
\newblock {\em Neural Networks 1}, 2, 119--130.

\bibitem[\protect\citename{Goodfellow et~al\mbox{.} }2014]{GoodfellowNIPS2014}
{\sc Goodfellow, I., Pouget-Abadie, J., Mirza, M., Xu, B., Warde-Farley, D.,
  Ozair, S., Courville, A., and Bengio, Y.}
\newblock 2014.
\newblock Generative adversarial nets.
\newblock In {\em Conference on Neural Information Processing Systems}.

\bibitem[\protect\citename{Grimm and Joshi
  }2012]{Grimm:2012:JDS:2331067.2331084}
{\sc Grimm, C., and Joshi, P.}
\newblock 2012.
\newblock Just drawit: A 3d sketching system.
\newblock In {\em nternational Symposium on Sketch-Based Interfaces and
  Modeling}, 121--130.

\bibitem[\protect\citename{Hilaire and Tombre }2006]{HilairePAMI2006}
{\sc Hilaire, X., and Tombre, K.}
\newblock 2006.
\newblock Robust and accurate vectorization of line drawings.
\newblock {\em IEEE Transactions on Pattern Analysis and Machine Intelligence
  28}, 6, 890--904.

\bibitem[\protect\citename{Igarashi et~al\mbox{.}
  }1997]{Igarashi:1997:IBT:263407.263525}
{\sc Igarashi, T., Matsuoka, S., Kawachiya, S., and Tanaka, H.}
\newblock 1997.
\newblock Interactive beautification: A technique for rapid geometric design.
\newblock In {\em ACM Symposium on User Interface Software and Technology},
  105--114.

\bibitem[\protect\citename{Iizuka et~al\mbox{.} }2016]{IizukaSIGGRAPH2016}
{\sc Iizuka, S., Simo-Serra, E., and Ishikawa, H.}
\newblock 2016.
\newblock {Let there be Color!: Joint End-to-end Learning of Global and Local
  Image Priors for Automatic Image Colorization with Simultaneous
  Classification}.
\newblock {\em ACM Transactions on Graphics (Proceedings of SIGGRAPH) 35}, 4.

\bibitem[\protect\citename{Ioffe and Szegedy }2015]{IoffeICML2015}
{\sc Ioffe, S., and Szegedy, C.}
\newblock 2015.
\newblock Batch normalization: Accelerating deep network training by reducing
  internal covariate shift.
\newblock In {\em International Conference on Machine Learning}.

\bibitem[\protect\citename{Isola et~al\mbox{.} }2016]{IsolaARXIV2016}
{\sc Isola, P., Zhu, J., Zhou, T., and Efros, A.~A.}
\newblock 2016.
\newblock Image-to-image translation with conditional adversarial networks.
\newblock {\em arXiv preprint arXiv:1611.07004\/}.

\bibitem[\protect\citename{Kang et~al\mbox{.} }2007]{KangNPAR2007}
{\sc Kang, H., Lee, S., and Chui, C.~K.}
\newblock 2007.
\newblock Coherent line drawing.
\newblock In {\em International Symposium on Non-Photorealistic Animation and
  Rendering}, 43--50.

\bibitem[\protect\citename{LeCun et~al\mbox{.} }1989]{LecunNM1989}
{\sc LeCun, Y., Boser, B., Denker, J.~S., Henderson, D., Howard, R.~E.,
  Hubbard, W., and Jackel, L.~D.}
\newblock 1989.
\newblock Backpropagation applied to handwritten zip code recognition.
\newblock {\em Neural computation 1}, 4, 541--551.

\bibitem[\protect\citename{Lindlbauer et~al\mbox{.} }2013]{LindlbauerHHSS2013}
{\sc Lindlbauer, D., Haller, M., Hancock, M.~S., Scott, S.~D., and
  Stuerzlinger, W.}
\newblock 2013.
\newblock Perceptual grouping: selection assistance for digital sketching.
\newblock In {\em International Conference on Interactive Tabletops and
  Surfaces}, 51--60.

\bibitem[\protect\citename{Liu et~al\mbox{.} }2015]{LiuSIGGRAPH2015}
{\sc Liu, X., Wong, T.-T., and Heng, P.-A.}
\newblock 2015.
\newblock Closure-aware sketch simplification.
\newblock {\em ACM Transactions on Graphics (Proceedings of SIGGRAPH Asia) 34},
  6, 168:1--168:10.

\bibitem[\protect\citename{Lu et~al\mbox{.} }2012]{LuNPAR2012}
{\sc Lu, C., Xu, L., and Jia, J.}
\newblock 2012.
\newblock Combining sketch and tone for pencil drawing production.
\newblock In {\em International Symposium on Non-Photorealistic Animation and
  Rendering}, 65--73.

\bibitem[\protect\citename{Mirza and Osindero }2014]{MirzaNIPSW2014}
{\sc Mirza, M., and Osindero, S.}
\newblock 2014.
\newblock Conditional generative adversarial nets.
\newblock In {\em Conference on Neural Image Processing Deep Learning
  Workshop}.

\bibitem[\protect\citename{Nair and Hinton }2010]{NairICML2010}
{\sc Nair, V., and Hinton, G.~E.}
\newblock 2010.
\newblock Rectified linear units improve restricted boltzmann machines.
\newblock In {\em International Conference on Machine Learning}, 807--814.

\bibitem[\protect\citename{Noh et~al\mbox{.} }2015]{NohICCV2015}
{\sc Noh, H., Hong, S., and Han, B.}
\newblock 2015.
\newblock Learning deconvolution network for semantic segmentation.
\newblock In {\em International Conference on Computer Vision}.

\bibitem[\protect\citename{Noris et~al\mbox{.} }2013]{NorisTOG2013}
{\sc Noris, G., Hornung, A., Sumner, R.~W., Simmons, M., and Gross, M.}
\newblock 2013.
\newblock Topology-driven vectorization of clean line drawings.
\newblock {\em ACM Transactions on Graphics 32}, 1, 4:1--4:11.

\bibitem[\protect\citename{Orbay and Kara }2011]{5710858}
{\sc Orbay, G., and Kara, L.}
\newblock 2011.
\newblock Beautification of design sketches using trainable stroke clustering
  and curve fitting.
\newblock {\em IEEE Transactions on Visualization and Computer Graphics 17}, 5,
  694--708.

\bibitem[\protect\citename{Radford et~al\mbox{.} }2016]{RadfordICLR2016}
{\sc Radford, A., Metz, L., and Chintala, S.}
\newblock 2016.
\newblock Unsupervised representation learning with deep convolutional
  generative adversarial networks.
\newblock In {\em International Conference on Learning Representations}.

\bibitem[\protect\citename{Rumelhart et~al\mbox{.} }1986]{RumelhartNATURE1986}
{\sc Rumelhart, D., Hinton, G., and Williams, R.}
\newblock 1986.
\newblock Learning representations by back-propagating errors.
\newblock In {\em Nature}.

\bibitem[\protect\citename{Salimans et~al\mbox{.} }2016]{SalimansNIPS2016}
{\sc Salimans, T., Goodfellow, I., Zaremba, W., Cheung, V., Radford, A., and
  Chen, X.}
\newblock 2016.
\newblock Improved techniques for training gans.
\newblock In {\em Conference on Neural Information Processing Systems}.

\bibitem[\protect\citename{Shesh and Chen }2008]{CGF:CGF1151}
{\sc Shesh, A., and Chen, B.}
\newblock 2008.
\newblock Efficient and dynamic simplification of line drawings.
\newblock {\em Computer Graphics Forum 27}, 2, 537--545.

\bibitem[\protect\citename{Simo-Serra et~al\mbox{.}
  }2016]{SimoSerraSIGGRAPH2016}
{\sc Simo-Serra, E., Iizuka, S., Sasaki, K., and Ishikawa, H.}
\newblock 2016.
\newblock {Learning to Simplify: Fully Convolutional Networks for Rough Sketch
  Cleanup}.
\newblock {\em ACM Transactions on Graphics (Proceedings of SIGGRAPH) 35}, 4.

\bibitem[\protect\citename{Srivastava et~al\mbox{.} }2014]{SrivastavaJMLR2014}
{\sc Srivastava, N., Hinton, G., Krizhevsky, A., Sutskever, I., and
  Salakhutdinov, R.}
\newblock 2014.
\newblock Dropout: A simple way to prevent neural networks from overfitting.
\newblock {\em Journal of Machine Learning Research 15\/}, 1929--1958.

\bibitem[\protect\citename{Zeiler }2012]{ZeilerARXIV2012}
{\sc Zeiler, M.~D.}
\newblock 2012.
\newblock {ADADELTA:} an adaptive learning rate method.
\newblock {\em arXiv preprint arXiv:1212.5701\/}.

\end{thebibliography}

\end{document}